%% file: main.tex
\documentclass[10pt,twocolumn]{article}

\usepackage[margin=0.75in]{geometry}
\usepackage[T1]{fontenc}
\usepackage{lmodern}
\usepackage{microtype}
\usepackage{times}

\usepackage{graphicx}
\usepackage{booktabs}
\usepackage{multirow}
\usepackage{amsmath}
\usepackage{amssymb}
\usepackage{siunitx}
\usepackage{makecell}
\usepackage{algorithm}
\usepackage{algpseudocode}
\usepackage{placeins}
\usepackage{xcolor}
\usepackage{enumitem}
\usepackage{url}
\usepackage[hidelinks]{hyperref}
\usepackage[nameinlink,noabbrev]{cleveref}

\title{BALLAST: Bandit-Assisted Learning for Latency-Aware Stable Timeouts in Raft}
\author{
Qizhi Wang\\
PingCAP, Data \& AI-Innovation Lab\\
Beijing, China\\
\texttt{qizhi.wang@pingcap.com}
}
\date{}

\begin{document}
\maketitle

\begin{abstract}
Randomized election timeouts are a simple and effective liveness heuristic for Raft, but they become brittle under long-tail latency, jitter, and partition recovery, where repeated split votes can inflate unavailability.
This paper presents BALLAST, a lightweight online adaptation mechanism that replaces static timeout heuristics with contextual bandits.
BALLAST selects from a discrete set of timeout ``arms'' using efficient linear contextual bandits (LinUCB variants), and augments learning with safe exploration to cap risk during unstable periods.
We evaluate BALLAST on a reproducible discrete-event simulation with long-tail delay, loss, correlated bursts, node heterogeneity, and partition/recovery turbulence.
Across challenging WAN regimes, BALLAST substantially reduces recovery time and unwritable time compared to standard randomized timeouts and common heuristics, while remaining competitive on stable LAN/WAN settings.
\end{abstract}

\section{Introduction}
\label{sec:intro}

Leader-based replication systems depend on timely leader recovery.
In Raft, leadership is re-established via leader election, whose liveness depends critically on election timeouts and failure detection.
The standard approach randomizes election timeouts within a fixed range to reduce split votes and avoid perpetual contention~\cite{ongaro2014raft}.
In modern deployments---geo-distributed clusters, multi-tenant clouds, and Kubernetes environments---network delays are often long-tailed and non-stationary~\cite{dean2013tail}.
Under these conditions, fixed timeout ranges can become miscalibrated: too aggressive leads to cascading split votes and term churn; too conservative inflates recovery latency and unwritable time.

\paragraph{Motivating observation.}
In production, operators often ``fix'' leader-election instability by pushing timeouts upward (e.g., from hundreds of milliseconds to seconds).
This can stop election storms, but it also turns every recovery event into a longer outage window.
Worse, the ``right'' timeout depends on where the cluster runs today (LAN vs.\ WAN), what the network looks like now (transient congestion), and even which nodes are slow (hardware or noisy neighbors).
As a result, timeout configuration becomes both high-stakes and environment-specific, and the same static setting can oscillate between being too aggressive and too conservative.

\textbf{Goal.} We seek an election-timeout mechanism that (i) adapts \emph{online} to changing network conditions, (ii) is lightweight enough to run in a consensus event loop, and (iii) includes safety valves that prevent catastrophic exploration during turbulence.

\textbf{Key idea.} We cast timeout selection as an online decision problem.
At each node and term, an agent observes local signals (e.g., heartbeat inter-arrival and election history) and selects a timeout \emph{arm} from a small discrete set (aggressive/moderate/conservative).
After an election attempt, the agent receives a delayed reward reflecting success and time-to-recover.
We instantiate this with linear contextual bandits (LinUCB)~\cite{li2010contextual} and add (a) non-stationary variants (discounted/sliding-window) and (b) safe exploration via a conservative fallback policy.

\paragraph{Where the idea comes from.}
BALLAST is inspired by a pragmatic view of consensus engineering: many ``hard'' availability incidents reduce to miscalibrated thresholds under shifting environments.
Failure detectors (e.g., \(\Phi\) accrual~\cite{hayashibara2004phi}) already embody this idea by adapting suspicion to observed arrival patterns.
We ask: can we apply a similarly lightweight, more directly \emph{optimization-driven} adaptation to election timeouts, while keeping the Raft protocol unchanged?

\textbf{Contributions.}
\begin{itemize}[leftmargin=*,itemsep=2pt]
  \item \textbf{BALLAST}, a lightweight contextual-bandit framework for Raft election timeouts with safe exploration and non-stationary adaptation.
  \item A reproducible evaluation methodology (discrete-event simulation, fault injection, protocol-level logging, CI-based aggregation) to study election stability under tail latency and recovery turbulence.
  \item An empirical study showing that BALLAST reduces recovery time and unwritable time relative to randomized timeouts and widely used heuristics, without sacrificing stable-LAN performance.
\end{itemize}

\section{Problem Setting and Metrics}
\label{sec:problem}

We focus on Raft leader election liveness (safety remains governed by the original protocol rules~\cite{ongaro2014raft}).
We report both election-process latency and end-to-end recovery.

\textbf{Election latency (process).} \texttt{time\_to\_leader} measures the duration from the start of a candidate election attempt to leader establishment.

\textbf{Recovery time (end-to-end).} \texttt{recovery\_time} measures the duration of an \emph{unwritable} interval, which includes waiting for election timeouts and retries.
We define \emph{writable} when a strict majority has recently observed heartbeats from the same leader within a grace window; otherwise the system is considered unwritable.
This captures the availability impact of split votes and partition recovery beyond the election process itself.
Unless otherwise stated, the grace window is set to
\(\texttt{heartbeat\_grace\_ms}=\max(3\cdot \texttt{heartbeat\_interval\_ms},\,2\cdot \texttt{tick\_ms})\)
as implemented in our metric computation, which corresponds to \SI{150}{ms} in the main scenario (\SI{50}{ms} heartbeats, \SI{10}{ms} ticks).
This window mirrors common Raft-style heartbeat expectations and avoids counting brief jitter as an unwritable period; we apply it uniformly across methods for comparability.

\textbf{Split-vote rate (stability).} We report \texttt{split\_vote\_rate} as the fraction of candidate elections that time out without a leader (an \texttt{election\_failed} event). This is a conservative proxy for split votes or quorum loss; we use it as a stability diagnostic rather than a protocol-level tie detector. We therefore avoid interpreting it as an exact split-vote count, and any ``split-vote'' references in the main text should be read as this proxy.
\textbf{Failure-cause breakdown (diagnostic).} To disambiguate the proxy, we also classify failed elections into (i) \emph{no quorum alive} (alive nodes $<$ majority), (ii) \emph{low reachability} (RequestVote observed by fewer than a majority of peers), and (iii) \emph{contention-like} (majority reachable but no leader). We report this breakdown in \Cref{sec:appendix-failure-breakdown} to contextualize proxy-driven churn.

\paragraph{Why these metrics.}
Election-latency metrics isolate the speed of a \emph{single} successful election attempt, but can miss repeated failures, cooldowns, and recovery turbulence.
In contrast, recovery time and unwritable fraction reflect the operator-facing question: ``how long did the system fail to accept writes?''.
Throughout, we treat recovery and unwritable time as the primary availability signals and use election-latency metrics to diagnose trade-offs.
\section{Background}
\label{sec:background}

\subsection{Raft election timeouts}
Raft elects a leader by timeouts and randomized elections: a follower becomes a candidate if it does not receive heartbeats before its election deadline, then solicits votes from a majority~\cite{ongaro2014raft}.
Randomization reduces the probability that multiple candidates time out simultaneously and split the vote.
However, the timeout range is typically static and tuned for expected network conditions, which can diverge substantially under long-tail or transient congestion.

\paragraph{Why static ranges are brittle.}
Election timeouts play a dual role: they are both a liveness parameter (how quickly we try to recover) and an implicit failure detector threshold (when we suspect the leader is down).
When the network is fast and stable, aggressive timeouts can reduce recovery latency.
Under congestion or long-tail delay, the same aggressiveness increases the probability that multiple nodes time out ``together'' from the perspective of delayed heartbeats, which raises split-vote probability and induces term churn.
In practice, operators often widen the timeout range to regain stability, but doing so can inflate every subsequent recovery event.

\subsection{Failure detection and tail latency}
Timeouts are a practical form of failure detection.
Classical theory formalizes liveness via unreliable failure detectors~\cite{chandra1996unreliable}, while systems often implement adaptive suspicion levels such as the $\Phi$ accrual failure detector~\cite{hayashibara2004phi}.
At scale, tail latency can dominate end-to-end behavior~\cite{dean2013tail}, making static thresholds brittle.

\paragraph{Heuristics vs.\ optimization.}
Adaptive failure detectors like \(\Phi\) accrual convert observed inter-arrival times into a suspicion level.
They are effective but still require choosing interpretation thresholds and may not directly optimize an availability objective such as unwritable time.
This motivates a learning-based approach that treats timeout selection as an online optimization problem with explicit objectives and measurable outcomes.

\subsection{Contextual bandits for online control}
Contextual bandits model online decision-making where an agent chooses an action given context and observes a reward, balancing exploration and exploitation.
LinUCB provides an efficient linear contextual bandit with an upper-confidence bound exploration bonus~\cite{li2010contextual}.
Unlike deep RL, LinUCB is computationally light and amenable to embedding in event loops.

\paragraph{Why bandits fit timeout selection.}
Timeout selection has three properties that align with contextual bandits:
(i) the action space is naturally small (a few reasonable timeout ranges),
(ii) feedback is delayed but frequent (every election attempt), and
(iii) the environment is partially observable and non-stationary (latency regimes shift).
Bandits offer a practical middle ground between fixed heuristics and full reinforcement learning: they can adapt online with small models while remaining interpretable and cheap to compute.
\section{Design: BALLAST}
\label{sec:design}

\subsection{Action space}
BALLAST selects from a small discrete set of timeout \emph{arms}.
Each arm corresponds to a timeout range $[T_{\min},T_{\max}]$, from which a node samples its election deadline:
\begin{itemize}[leftmargin=*,itemsep=2pt]
  \item \textbf{A1 (aggressive)}: \SIrange{150}{300}{ms}
  \item \textbf{A2 (moderate)}: \SIrange{300}{600}{ms}
  \item \textbf{A3 (conservative)}: \SIrange{600}{1200}{ms}
\end{itemize}
Sampling within an arm preserves Raft's symmetry-breaking randomization~\cite{ongaro2014raft} while allowing the agent to move the \emph{range} to match the current latency regime.
We intentionally keep the action space small to limit risk and to ease interpretability: operators can reason about the meaning of each arm (``fast'', ``medium'', ``safe'').
We also evaluate broader and shifted arm sets, plus quantile-anchored arm construction, as sensitivity studies in \Cref{sec:appendix-arms}.
\paragraph{Data-driven arm scaling.}
To reduce dependence on handpicked absolute ranges, we also evaluate a simple arm-construction rule that derives a \emph{base} timeout from an online quantile of observed heartbeat delays and defines arms as relative multiplier ranges (e.g., \([3\!-\!5]\times, [5\!-\!7]\times, [7\!-\!9]\times\)).
This yields \texttt{bandit\_qdecay}, which combines quantile-based scaling with bandit selection over relative arms.

\subsection{Context features}
The context vector $x_t$ uses local, inexpensive signals available at each node (no global coordination):
\begin{itemize}[leftmargin=*,itemsep=2pt]
  \item heartbeat inter-arrival statistics (mean/std) and time since last heartbeat,
  \item recent election outcomes (consecutive failures),
  \item a regime identifier (optional) for privileged-information oracle baselines.
\end{itemize}
These features are intentionally small-dimensional to keep updates and inference lightweight.
\paragraph{Design principle.}
We favor features that are (i) locally observable in typical Raft implementations, (ii) robust to transient noise, and (iii) directly connected to liveness failure modes (missed heartbeats, repeated failures).
This is important for deployability: the policy should not require expensive instrumentation or cluster-wide telemetry to make safe decisions.
\paragraph{Feature scaling.}
We feed raw millisecond-valued features (plus a bias) without per-feature normalization or per-arm scaling.
Ridge regularization (\(\ell_2\)) and a small action space keep updates stable in practice; feature-set ablations are reported in \Cref{sec:ablation}.
Exploring alternative normalization or adaptive scaling is left to future work.

\subsection{Reward}
Rewards should prefer successful recovery while discouraging slow and unstable elections.
We use a shaped reward
\[
r = w_s \cdot \mathbb{I}[\text{success}] - w_\ell \cdot \text{latency} - w_{sv} \cdot \mathbb{I}[\text{split vote}],
\]
where latency is measured in milliseconds and coefficients are configurable.
The reward is \emph{delayed}: it is observed only after an election attempt completes (leader elected) or times out and fails.
\paragraph{Interpretation.}
The success term rewards restoring leadership; the latency penalty discourages trivially conservative timeouts; the split-vote penalty discourages unstable ``thrashing'' where many nodes campaign and no leader emerges.
In our experiments, we emphasize end-to-end recovery (unwritable intervals), and use the reward primarily as a learning signal to navigate the trade-off between speed and stability.
\paragraph{Weighting rationale.}
We set \(w_s{=}1\) and \(w_\ell{=}0.002\) so that roughly \SI{500}{ms} of extra latency offsets one unit of success reward, and use \(w_{sv}{=}1\) to penalize failed elections comparably to a miss.
Sensitivity to these weights is reported in \Cref{sec:appendix-tuning-sensitivity}.
\paragraph{Operational guidance.}
A simple way to align weights with SLOs is to choose \(w_\ell \approx 1/\Delta T\), where \(\Delta T\) is the additional recovery latency (in ms) an operator would trade for one successful election; e.g., \(\Delta T{=}500\) yields \(w_\ell{=}0.002\).
\(w_{sv}\) sets the penalty of a failed election relative to a success; \(w_{sv}\in[0.5,2]\) keeps this penalty on the same order as \(w_s\) while allowing stability- or speed-oriented tuning.
Larger \(w_{sv}\) (or smaller \(F\) in the safety gate) biases toward lower unwritable fraction, while smaller \(w_{sv}\) (or larger \(F\)) biases toward faster recoveries.
Our sensitivity study in \Cref{sec:appendix-tuning-sensitivity} provides empirical trade-offs across these ranges.
\paragraph{Split-vote proxy.}
In the simulator, we set \texttt{split\_vote}{=}1 when a candidate election times out without reaching a majority (\texttt{election\_failed}); this is a conservative proxy for split votes or quorum loss, and is used only as a churn penalty in the learning signal.
This proxy does not perfectly separate contention from transient quorum loss; we therefore report a post-hoc breakdown based on alive-quorum and RequestVote reachability in \Cref{sec:appendix-failure-breakdown}.

\subsection{Learning: LinUCB with non-stationarity}
We employ LinUCB~\cite{li2010contextual}, which estimates linear payoffs per arm and uses an upper-confidence bound for exploration.
To handle regime changes, BALLAST supports:
\begin{itemize}[leftmargin=*,itemsep=2pt]
  \item \textbf{Discounted LinUCB}: exponentially down-weights older observations,
  \item \textbf{Sliding-window LinUCB}: keeps a bounded recent history.
\end{itemize}
These are standard techniques for non-stationary bandits~\cite{russac2019dlinucb,chen2019nonstationary}.
\paragraph{Why linear.}
LinUCB is a pragmatic choice: it is easy to implement safely, provides a clear exploration mechanism, and can be updated with a few linear-algebra operations.
This supports the systems requirement that decisions should not add noticeable overhead to the event loop.

\subsection{Safe exploration}
Exploration can be dangerous: choosing an overly aggressive arm during turbulence can trigger election storms.
BALLAST wraps the learner with a conservative safety layer: after $F$ consecutive failures, the policy enters a cooldown period where it forces a safe arm (e.g., A3).
We log safety triggers and provide their frequency in the artifact to quantify risk-control behavior.
\paragraph{Concrete parameters and re-entry.}
In our default configuration, $F{=}3$ and \texttt{cooldown\_elections}{=}2, and the safe arm is the most conservative arm in the current action set (A3 for the 3-arm setting; A5 for the 5-arm sensitivity setting).
Cooldown is entered when a candidate times out and the local consecutive-failure counter reaches $F$.
During cooldown, the policy forces the safe arm on subsequent deadline resets.
Cooldown is reduced by one after each successful election; otherwise it persists.
\paragraph{Why this safety model.}
The safety wrapper is deliberately simple and conservative.
It enforces a human-understandable rule: ``if elections keep failing, stop being aggressive and wait longer''.
This is not a formal safety proof for distributed liveness, but it provides an operationally meaningful guardrail that limits downside risk during unstable periods.

\subsection{Algorithm sketch}
\Cref{alg:ballast} summarizes the control loop at a single node.
Each node maintains its own lightweight model; no node-to-node learning coordination is required.

\begin{algorithm}[htbp]
\caption{BALLAST at one node (per-election update)}
\label{alg:ballast}
\begin{algorithmic}[1]
\State Initialize per-arm parameters \(\{A_a, b_a\}\) (LinUCB)
\For{each election attempt \(t\)}
  \State Observe context \(x_t\) from local heartbeat/election history
  \If{safety triggered (e.g., \(F\) consecutive failures)}
    \State Choose safe arm \(a_t \leftarrow \text{A3}\)
  \Else
    \State Choose arm \(a_t \leftarrow \arg\max_a \text{UCB}_a(x_t)\)
  \EndIf
  \State Sample timeout \(T_t \sim \text{Uniform}(T_{\min}(a_t), T_{\max}(a_t))\)
  \State Run election with deadline \(T_t\); observe outcome and latency
  \State Compute reward \(r_t\) and update only the chosen arm \((A_{a_t}, b_{a_t})\)
\EndFor
\end{algorithmic}
\end{algorithm}
\section{Experimental Methodology}
\label{sec:impl}

\textbf{Goal.}
Our evaluation goal is not to ``beat'' Raft's safety (which remains unchanged), but to quantify how timeout policies affect \emph{liveness and availability} under modern network conditions: long-tail delay, non-stationary jitter, correlated loss, partitions, and heterogeneous node speeds.
These conditions are widely reported in large-scale deployments and are difficult to reproduce deterministically in a live cluster.

\textbf{Why discrete-event simulation.}
Leader election is a timing feedback loop: timeouts drive candidacy, candidacy drives network load, and network load feeds back into timeouts via delayed heartbeats and vote RPCs.
Under long-tail latency, small parameter changes can induce qualitatively different behaviors (e.g., stable leadership vs.\ election storms).
A discrete-event simulator provides two key advantages:
(i) \emph{control} (we can inject specific failure/latency regimes and repeat them exactly), and
(ii) \emph{observability} (we can log protocol-level events and derive metrics consistently across policies).
We emphasize that the simulator models the Raft election and heartbeat mechanisms at message granularity with configurable link delay/loss processes; it is designed to stress liveness, not to emulate a full storage engine.

\textbf{Network and fault injection.}
Each scenario specifies a link model (e.g., stable WAN vs.\ long-tail delay), optional regime switches (to emulate non-stationarity), and fault episodes (crash/restart, partitions with recovery turbulence).
We include correlated burst loss because it is a common source of transient failure suspicion.
Node heterogeneity is modeled by per-node service delays to capture slow followers.

\textbf{Policies and measurement.}
All compared policies run on identical seeds and injected events.
For each run, we record protocol events (timeouts, candidacy, votes, leader establishment, heartbeats) and then derive liveness/availability metrics offline.
This separation avoids ``baking in'' policy-specific instrumentation and makes ablations comparable.
For learning-based policies, we also log per-election outcomes (success/failure, latency, and split-vote timeouts) so that the shaped reward in \Cref{sec:design} can be computed and applied consistently.
We additionally log RequestVote receipts and node crash/restart events to classify failed elections into quorum-loss vs.\ contention-like causes (reported in \Cref{sec:appendix-failure-breakdown}).

\paragraph{Cold start and state persistence.}
Bandit models initialize with a ridge prior (\(A{=}\lambda I, b{=}0\), where \(\lambda\) is the \(\ell_2\) regularization coefficient) and no explicit burn-in; ties in LinUCB are broken by arm index, so early choices can skew toward the first arm until data accrue.
In the simulator, crash/restart events remove nodes from the schedule but do not reset in-memory policy state by default; in real deployments, a process restart would reset the learner unless state is persisted.
We therefore include a reset-on-restart variant that clears the policy state at restart to emulate cold starts (\Cref{sec:appendix-restart}), and find that the safety wrapper largely stabilizes early behavior.
Persisting a small bandit snapshot (per-node \(A,b\) matrices and wrapper state) to stable storage is straightforward and left for future implementation in production systems.
In practice, persisting on major transitions (e.g., after a successful election) or on a coarse cadence (e.g., every 10--50 election outcomes) keeps write-amplification low while bounding cold-start risk during rolling upgrades.

\textbf{Statistical reporting.}
We report point estimates aggregated across seeds and 95\% bootstrap confidence intervals (percentile bootstrap over seeds).
This is a lightweight, assumption-minimal way to convey variability without requiring distributional assumptions about recovery times.
\paragraph{Artifact availability.}
We will release reproducible code, configurations, and machine-readable data artifacts (per-seed logs, derived metrics, and confidence intervals) at \url{https://github.com/Icemap/ballast}.
\section{Evaluation}
\label{sec:eval}

\subsection{Experimental setup}
We evaluate four scenarios:
\textbf{(i)} a hard long-tail WAN with crash/restart, partition, and regime switch (main),
\textbf{(ii)} partition recovery with post-recovery turbulence,
\textbf{(iii)} stable WAN with moderate delay/jitter,
and \textbf{(iv)} a LAN no-regression sanity check.
Unless otherwise stated, each scenario runs a $N\in\{5,7,9\}$ node cluster for \SI{60}{s} simulated time (LAN uses \SI{30}{s}) with \SI{50}{ms} heartbeats.
We report point estimates aggregated over seeds with 95\% bootstrap confidence intervals (CI) (percentile bootstrap over seeds).
\paragraph{Reproducibility.}
Each method is evaluated on exactly the same injected scenarios and random seeds, ensuring that differences are attributable to the timeout policy rather than to different fault realizations.
We include confidence intervals to reflect run-to-run variability in election dynamics, which can be high under long-tail latency and partitions.

\paragraph{Baselines.}
We compare against common timeout policies: \texttt{random} (standard Raft randomized range), \texttt{static\_conservative}, \texttt{backoff}, \texttt{rtt\_heuristic}, and \texttt{phi\_accrual} (\(\Phi\) accrual failure detection).
We additionally include \texttt{quantile\_decay}, a simple adaptive baseline that expands/contracts the timeout range using a decayed online quantile estimate from observed heartbeat delays; and \texttt{bandit\_qdecay}, which uses the same quantile estimator to set a \emph{moving} timeout scale while a contextual bandit selects among a small set of relative multiplier arms (reducing dependence on handpicked absolute ranges).
We evaluate two Dynatune-inspired baselines: \texttt{dynatune\_et}, an election-timeout-only variant that computes an RTT/one-way-delay-based timeout and keeps heartbeat intervals fixed, and \texttt{dynatune\_joint}, a simplified co-tuning variant that adjusts both heartbeat interval and election timeout using the same base.
We also include \texttt{bandit\_ts\_safe}, a non-stationary Thompson-sampling variant (discounted linear TS) with the same safety wrapper as BALLAST.
BALLAST is \texttt{bandit\_safe}.
\texttt{oracle\_hint} (implemented as \texttt{oracle\_regime} in code) reveals the injected regime identifier to the policy and uses a simple hand-chosen regime$\rightarrow$arm mapping; it is a privileged-information baseline, \emph{not} a true upper bound.
Our Dynatune baselines are intentionally scoped: we do not implement per-follower timers or all measurement rules in the original system, and instead use a shared timeout/heartbeat base with explicit clamping to keep the comparison within our event-loop model.
For \texttt{dynatune\_et}, we follow Dynatune's RTT-based timeout logic but (i) fall back to one-way heartbeat delays when RTTs are unavailable, and (ii) enforce a minimum timeout-to-heartbeat ratio to avoid pathological aggressiveness; this yields an election-timeout-only comparison consistent with our problem scope.
We tune the Dynatune parameters (safety factor, minimum ratio, and clamping) per scenario using disjoint seeds; the selected values are reported in \Cref{tab:baseline-tuning}.
To provide a meaningful per-regime reference, we also report \texttt{oracle\_best\_per\_regime}, which enumerates regime$\rightarrow$arm mappings over the same discrete arm set and selects the best mapping on disjoint tuning seeds.
Dynatune is closely related; our \texttt{dynatune\_joint} baseline is a lightweight co-tuning heuristic intended to capture the qualitative benefit of joint tuning without reproducing the full system, while \texttt{dynatune\_et} remains a scoped election-timeout-only comparison~\cite{shiozaki2025dynatune}.
\paragraph{Baseline tuning protocol.}
For baselines with free thresholds (e.g., \texttt{rtt\_heuristic}, \texttt{phi\_accrual}, \texttt{quantile\_decay}, \texttt{bandit\_qdecay}, \texttt{bandit\_ts\_safe}, \texttt{dynatune\_et}, and \texttt{dynatune\_joint}), we tune parameters per scenario on a disjoint set of tuning seeds, then report results on the fixed report seeds.
The tuning objective is lexicographic: minimize unwritable fraction first, then recovery mean time.
For \texttt{oracle\_best\_per\_regime}, we enumerate all regime$\rightarrow$arm mappings over the discrete arm set (\(|\mathcal{A}|^{R}\) candidates for \(R\) regimes) on the tuning seeds and select the best mapping before evaluation on report seeds.
Unless otherwise stated, \texttt{bandit\_safe} uses the shaped reward in \Cref{sec:design} with coefficients \(w_s{=}1\), \(w_\ell{=}0.002\), and \(w_{sv}{=}1\), plus the conservative safety wrapper.
Unless otherwise stated, the bandit hyperparameters are \(\alpha{=}1.0\), \(l2{=}1.0\), and discount \(0.98\); the feature set is a 4-dimensional local statistic vector (heartbeat or RTT mean/std, time since last heartbeat, and consecutive failures), and safety uses \(F{=}3\), \texttt{cooldown\_elections}{=}2.

\subsection{Main results: hard WAN with tail latency}
\label{sec:eval-main}

\begin{figure}[htbp]
  \centering
  \includegraphics[width=\linewidth]{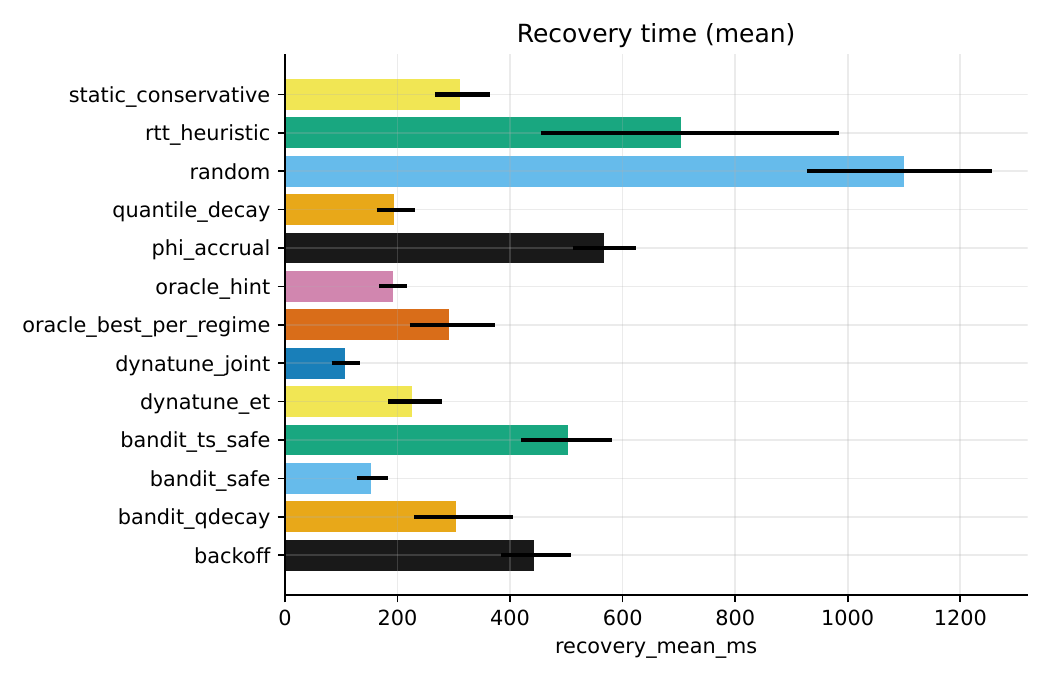}
  \caption{Main scenario (30 seeds): mean recovery time with 95\% CI. Lower is better.}
  \label{fig:main-recovery}
\end{figure}

\Cref{fig:main-recovery} and \Cref{tab:main-methods} summarize the main scenario over 30 seeds.
BALLAST significantly reduces end-to-end recovery time compared to randomized election timeouts and common heuristics, while remaining competitive on unwritable time.
Dynatune-inspired baselines are strong in this scenario: \texttt{dynatune\_joint} attains the fastest recovery, while \texttt{dynatune\_et} attains the lowest unwritable fraction.
Compared to conservative static timeouts, BALLAST improves recovery without paying the steady penalty of always waiting longer; compared to aggressive heuristics, it reduces split-vote-driven churn by escalating conservatively only when the local signals indicate turbulence.
\paragraph{Takeaway.}
In the most adversarial setting (tail latency, crash/restart, partition, and a mid-run regime switch), an adaptive timeout policy with explicit safety gating can materially reduce availability loss, even though it does not necessarily minimize conditional election latency on every successful attempt.

\begin{table}[htbp]
  \centering
  \scriptsize
  \setlength{\tabcolsep}{3pt}
  \resizebox{\linewidth}{!}{\input{tables/table_main_methods}}
\caption{Main scenario (30 seeds): recovery mean/p95/p99/max and unwritable fraction (95\% CI).}
  \label{tab:main-methods}
\end{table}

\begin{figure}[htbp]
  \centering
  \includegraphics[width=\linewidth]{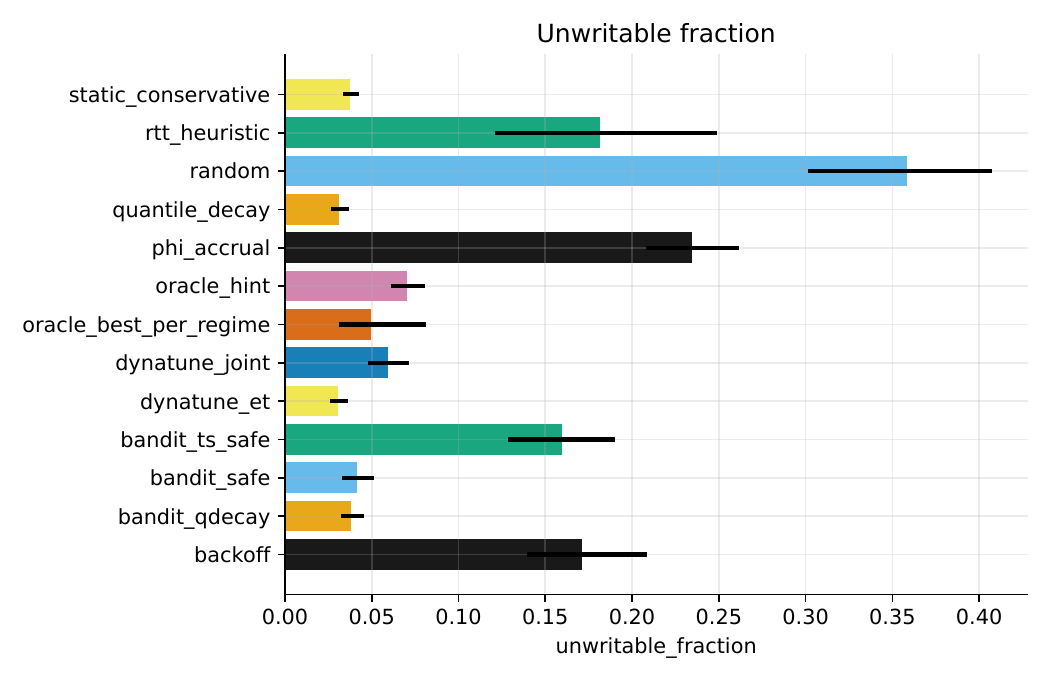}
  \caption{Main scenario (30 seeds): unwritable fraction with 95\% CI. Lower is better.}
  \label{fig:main-unwritable}
\end{figure}

\paragraph{Interpreting the trade-off.}
\Cref{fig:main-recovery,fig:main-unwritable} illustrates our primary optimization goal:
reduce end-to-end recovery time substantially while keeping unwritable time competitive.
In this scenario, \texttt{dynatune\_et} achieves the lowest unwritable fraction, while \texttt{dynatune\_joint} achieves the fastest recovery but with a higher unwritable fraction.
\texttt{quantile\_decay} is close to \texttt{dynatune\_et} on unwritable time but pays a recovery penalty; \texttt{bandit\_qdecay} narrows this gap by anchoring the arm scale to the quantile estimator while allowing bandit selection among relative ranges.
\texttt{static\_conservative} is also competitive on unwritable time but is substantially slower to recover.
BALLAST targets a practical balance: it maintains low unwritable time while improving recovery markedly relative to non-adaptive heuristics.

\paragraph{Tail recovery behavior.}
To align with operator concerns about extreme outages, we additionally report recovery p99 and per-seed worst-case recovery durations.
The tail metrics (including worst observed per-seed maxima) follow the same ordering as the means in the main scenario; see \Cref{tab:main-worstcase} in the appendix.

\begin{figure}[htbp]
  \centering
  \includegraphics[width=\linewidth]{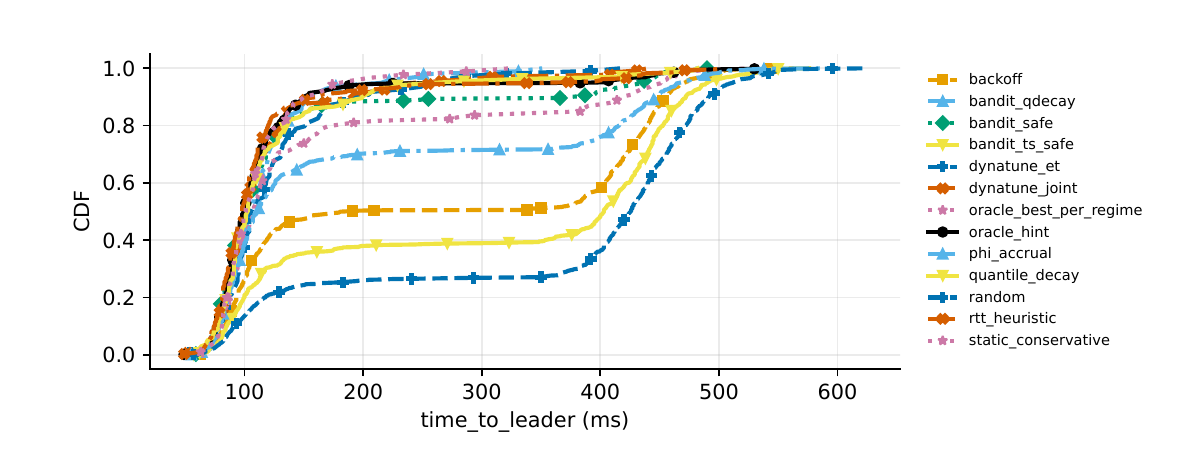}
  \caption{Main scenario: CDF of conditional election latency (\texttt{time\_to\_leader}).}
  \label{fig:main-cdf-ttl}
\end{figure}

\paragraph{Why election-latency CDFs can look ``mid''.}
Figure-level comparisons sometimes focus on conditional election-process latency (\texttt{time\_to\_leader}).
\Cref{fig:main-cdf-ttl} shows that \texttt{bandit\_safe} can sit between aggressive heuristics and conservative ones on this metric.
This is expected and, importantly, not contradictory to improved availability.
The safety layer occasionally forces conservative arms after consecutive failures, which can slow a \emph{single} election attempt.
However, by reducing repeated split votes and term churn, BALLAST shortens the \emph{overall} unwritable interval that an operator experiences.
In other words, ``slightly slower elections, fewer retries'' can beat ``fast elections, many failures'' on end-to-end recovery.
\paragraph{Dynatune ET-only vs.\ joint comparison.}
We report both \texttt{dynatune\_et} (timeout-only) and \texttt{dynatune\_joint} (heartbeat+timeout co-tuning) to reflect Dynatune's emphasis on joint parameter adjustment.
The ET-only variant can still lag under turbulence because fixed heartbeats limit failure-detection adaptivity; the joint variant generally improves over ET-only in our emulation, though it remains a simplified approximation rather than a full Dynatune reproduction.
We therefore interpret these results as scoped baselines within our event-loop model and emphasize that a full, protocol-faithful Dynatune replication is future work.

\subsection{Real Raft prototype (etcd/raft)}
\label{sec:eval-etcd}
To address the simulation-only limitation, we built a minimal \emph{real} prototype on \texttt{etcd/raft} (v3.5.x) in Go.
Each node runs the BALLAST policy in-process and selects a fresh election timeout whenever the Raft election timer is reset.
We deploy a 3-node cluster in Docker on a single host and inject network delay/loss via Linux \texttt{tc netem}; partitions and crash/restart events are scripted with \texttt{iptables} and container stop/start.
This is a \emph{single-host emulation}, not a multi-host production deployment, and is intended as a feasibility check under real networking/IPC and OS scheduling noise.
The prototype implements a focused policy subset (random/static/backoff/BALLAST), keeps membership fixed, and disables PreVote in the 3-node baseline; it does not model full storage workloads or log-replication backpressure.
We later add PreVote and a lightweight workload in an extended experiment (see \Cref{sec:appendix-etcd-prevote}).
We mirror the simulation scenarios (hard WAN with crash+partition+regime switch, partition recovery turbulence, stable WAN, and LAN) but scale to 3 nodes, running 10 seeds per policy.
Metrics use the same definitions as in the simulator (recovery time and unwritable fraction with the same heartbeat grace window), and we emphasize relative comparisons over absolute times.

\begin{figure}[htbp]
  \centering
  \includegraphics[width=\linewidth]{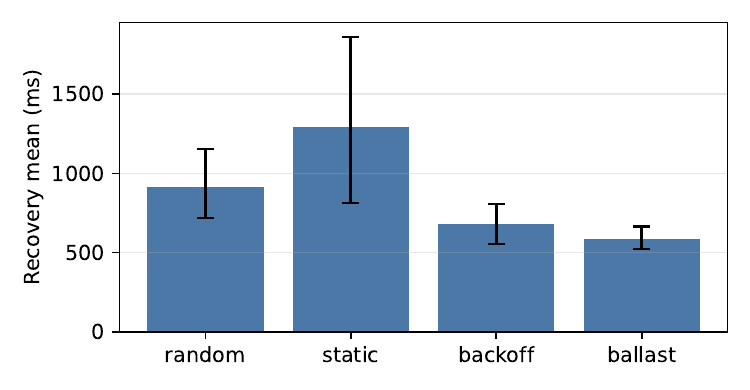}
  \caption{Real \texttt{etcd/raft} prototype: mean recovery time (95\% CI), main scenario.}
  \label{fig:etcd-raft-recovery}
\end{figure}

\begin{figure}[htbp]
  \centering
  \includegraphics[width=\linewidth]{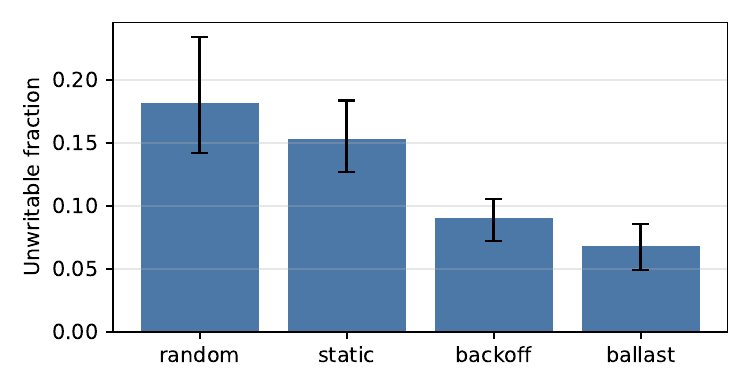}
  \caption{Real \texttt{etcd/raft} prototype: unwritable fraction (95\% CI), main scenario.}
  \label{fig:etcd-raft-unwritable}
\end{figure}

\paragraph{Main-scenario results.}
\Cref{tab:etcd-raft-main,fig:etcd-raft-recovery,fig:etcd-raft-unwritable} report the hard-WAN scenario on the real prototype.
Across 10 seeds, BALLAST reduces recovery time relative to randomized and backoff baselines and keeps unwritable fraction low; the direction matches the simulator, with conservative fallback limiting churn during turbulence.
The gains are measured in a minimal single-host emulation, so we emphasize directionality rather than production-scale claims.
\paragraph{Multi-AZ (no-netem) validation.}
To address the ``single-host'' limitation, we additionally ran the real prototype on three EC2 instances in AWS \texttt{ap-southeast-1}, spanning two availability zones (two nodes in \texttt{apse1-az3} and one in \texttt{apse1-az1}), under real network latencies only (no artificial shaping; \texttt{tc netem} disabled).
This is a normal network setting rather than BALLAST's most favorable stress regime.
Across 10 seeds of the main scenario, all methods are close and within overlapping CIs (\Cref{tab:etcd-raft-multi-az}); BALLAST remains competitive but is not consistently best.
We view this as a sanity check that the policy does not degrade under a real multi-AZ deployment with natural latencies.
\paragraph{PreVote + workload at larger $N$ (Docker+netem).}
To address concerns about PreVote and replication pressure, we extend the single-host Docker+netem prototype to $N{=}5$ and $N{=}7$ with PreVote enabled and a lightweight client workload (one proposal every \SI{50}{ms}, 256\,B payload) in the main scenario.
Results are summarized in \Cref{tab:etcd-raft-main-prevote-n5,tab:etcd-raft-main-prevote-n7}: BALLAST remains competitive with randomized and backoff baselines under the added election protocol phase and log replication traffic, and does not introduce clear regressions.
\paragraph{Host-level stress.}
To probe robustness under system noise, we additionally apply CPU+IO stress on each host during the same 3-node scenarios.
Results remain within overlapping CIs and show no clear regressions for BALLAST under stress (\Cref{tab:etcd-raft-stress,tab:etcd-raft-multi-az-stress}).

\begin{table}[htbp]
  \centering
  \scriptsize
  \resizebox{\linewidth}{!}{\input{tables/table_etcd_raft_multi_az_no_netem}}
  \caption{Real \texttt{etcd/raft} prototype on AWS multi-AZ (no \texttt{netem}): main scenario summary (95\% CI, 10 seeds).}
  \label{tab:etcd-raft-multi-az}
\end{table}

\begin{table}[htbp]
  \centering
  \scriptsize
  \resizebox{\linewidth}{!}{\input{tables/table_etcd_raft_main}}
  \caption{Real \texttt{etcd/raft} prototype (3-node Docker+netem): main scenario summary (95\% CI, 10 seeds).}
  \label{tab:etcd-raft-main}
\end{table}

\subsection{Additional scenarios (appendix)}
\label{sec:eval-more}
We evaluate additional settings (motivation regime shift, partition recovery turbulence, stable WAN, and LAN no-regression) to sanity-check robustness beyond the main stress test.
Because some of these scenarios produce small absolute differences where overlaid CDFs can be visually crowded, we place the full set of plots in \Cref{sec:appendix}.

\subsection{Overhead}
We microbenchmark the bandit policy in isolation to estimate per-decision and per-update overhead.
Across \num{50000} iterations, LinUCB selection and update remain lightweight, supporting integration into event loops.
We additionally instrument the etcd/raft prototype to record policy choose/update times under the emulated main scenario; p99 choose/update latency remains in the tens-of-microseconds range, with rare millisecond-scale outliers attributable to runtime scheduling/GC noise on a single host (\Cref{tab:etcd-overhead}).

\begin{table}[htbp]
  \centering
  \small
  \resizebox{\linewidth}{!}{\input{tables/table_overhead}}
  \caption{Policy overhead microbenchmark (single thread).}
  \label{tab:overhead}
\end{table}

\begin{table}[htbp]
  \centering
  \small
  \resizebox{\linewidth}{!}{\input{tables/table_etcd_overhead}}
  \caption{Policy overhead measured inside the etcd/raft prototype (single-host emulation).}
  \label{tab:etcd-overhead}
\end{table}
\section{Ablation}
\label{sec:ablation}

We perform ablations on a recovery-focused scenario with repeated partitions and regime switches.
The goal is to identify which design choices matter under recovery turbulence: safety gating, non-stationarity handling, exploration strength, reward shaping, and feature choices.

\paragraph{Self-check ablation protocol.}
A common pitfall in systems ablations is ``no-op'' variants where a knob change does not alter behavior under the chosen workload (e.g., the policy never enters the code path being ablated).
To avoid overstating conclusions, we add a self-check:
for each ablation variant, we compare per-seed event traces against the base policy; if all traces are identical, we mark the variant as invalid for this scenario.
We report only valid ablations in \Cref{tab:ablation} and list invalid variants in the appendix for transparency.

Key findings include:
\begin{itemize}[leftmargin=*,itemsep=2pt]
  \item \textbf{Safe exploration trades speed for stability:} removing the safety gate increases split-vote rate (more instability) even when mean recovery can look slightly better, reflecting that the wrapper is primarily a risk-control mechanism rather than a pure latency optimizer.
  \item \textbf{Non-stationarity handling matters:} sliding-window adaptation can improve recovery under regime switches by emphasizing recent network conditions; in this ablation scenario, it outperforms the base configuration.
  \item \textbf{Exploration strength affects stability:} changing the LinUCB exploration coefficient \(\alpha_{\text{ucb}}\) shifts the trade-off between faster adaptation and increased split votes; higher exploration improves recovery in this scenario but may raise risk under other regimes.
  \item \textbf{Reward shaping influences the trade-off:} a success-only reward can further reduce recovery time in this scenario, but it weakens explicit churn penalties that help guard against instability in other settings.
  \item \textbf{Feature scaling is not critical in our settings:} simple online z-score normalization yields comparable results, suggesting that raw feature magnitudes are not a dominant confounder in our scenarios.
\end{itemize}

\begin{table}[htbp]
  \centering
  \scriptsize
  \setlength{\tabcolsep}{3pt}
  \resizebox{\linewidth}{!}{\input{tables/table_ablation}}
  \caption{Self-check ablation: deltas relative to the base policy (mean over seeds). Negative is better for recovery and unwritable.}
  \label{tab:ablation}
\end{table}

All ablation variants share the same scenario and seeds as the base, differing only in the ablated component.
The base configuration is selected to provide a conservative, balanced trade-off across scenarios; some ablation variants outperform it in this specific turbulence setting, underscoring that optimal parameters are workload-dependent.
To make this trade-off explicit, we summarize cross-scenario deltas for the strongest ablation variants on main/partition/WAN/LAN in \Cref{tab:appendix-ablation-cross}.
We release the full ablation artifacts (event traces and derived metrics) to support independent verification and follow-up analyses.
\section{Discussion: Research Questions, Scope, and Validity}
\label{sec:discussion}

\subsection{Research questions and hypotheses}
This work studies a narrow but operationally important control point in consensus: election-timeout selection for Raft leader recovery.
We organize the study around three research questions.

\textbf{RQ1 (Availability under turbulence).}
Can a lightweight, online-learned timeout policy reduce end-to-end unavailability under long-tail delay, loss bursts, and partition recovery compared to commonly deployed heuristics?
\textbf{Hypothesis:} contextual adaptation can reduce repeated split votes and term churn, thereby shortening unwritable intervals.

\textbf{RQ2 (No-regression under stable regimes).}
Does online adaptation harm common steady-state deployments (stable WAN/LAN), where traditional randomized timeouts typically work well?
\textbf{Hypothesis:} with a small action space and conservative safety gating, the learned policy converges to competitive arms and avoids regressions.

\textbf{RQ3 (Deployability and operational risk).}
Can the learning component be made lightweight and predictable enough for systems engineers to accept in the consensus loop?
\textbf{Hypothesis:} a white-box linear contextual bandit plus explicit safety rules provides a favorable trade-off between adaptability, interpretability, and overhead.

\subsection{Contributions and boundaries}
BALLAST is intentionally scoped.
It does \emph{not} modify Raft safety rules, message formats, or quorum requirements; it only changes how election deadlines are chosen.
Accordingly, our contributions are about \emph{liveness/availability behavior} and \emph{operational simplicity}, not about new consensus correctness results.

We also focus on leader election and availability proxies rather than application-level throughput.
The interaction between election timing and full database stacks (replication pipelines, disk stalls, admission control) is important and left to system-specific validation.
Our evaluation methodology is designed to isolate timing feedback loops and to enable controlled ablations; it is not intended as a complete performance model of any particular database.
We do not provide formal convergence or safety guarantees for the learning dynamics under multi-agent coupling; the safety wrapper is an operational guardrail validated empirically in our scenarios.
\paragraph{Discrete arms vs.\ continuous tuning.}
We intentionally restrict the action space to a small set of interpretable timeout ranges; this limits risk and simplifies operator reasoning, but may underfit environments where a continuous or finer-grained timeout is optimal.
Exploring continuous action bandits or adaptive range expansion is a natural extension.
\paragraph{Co-tuning heartbeat intervals.}
We fix heartbeats to isolate election-timeout effects; co-tuning heartbeat cadence (as in Dynatune) could further reduce detection latency but introduces a second feedback loop.
Studying joint stability and the interaction between heartbeat cadence and timeout adaptation is future work.

\subsection{Threats to validity}
\textbf{Construct validity.}
Availability is multi-dimensional.
We use recovery time and unwritable fraction as primary metrics because they reflect operator-visible write unavailability, but they do not capture all user-facing SLOs (e.g., tail read latency).
We mitigate this by also reporting election stability metrics (split-vote rate, term churn) that help interpret why recovery improves or degrades.

\textbf{Internal validity.}
Learning systems can be sensitive to reward shaping and feature design.
We address this with self-check ablations and non-stationary variants to verify that the main claims are not a fragile consequence of a single tuning choice.
We further add sensitivity to reward weights, safety thresholds, and feature normalization, but we do not claim exhaustive production guidance; alternative reward formulations could still shift the speed--stability trade-off.

\textbf{External validity.}
Most results are obtained in a controlled discrete-event setting.
We also validate a minimal \texttt{etcd/raft} prototype under Docker+\texttt{netem} emulation (\Cref{sec:eval-etcd}), which introduces OS scheduling noise and real networking/IPC overheads.
However, real deployments have additional sources of variance: GC pauses, storage stalls, and application backpressure.
Deployment studies in production-grade Raft implementations and full database stacks remain necessary future work.
Our prototype keeps membership fixed; while we add PreVote in a limited Docker+netem extension, we only approximate GC/IO contention via host-level stress and do not exercise full client workloads.
We do not model membership changes; conservative cooldowns could, in principle, delay legitimate leader transitions during reconfiguration, which warrants targeted validation.
\paragraph{Protocol-level alternatives.}
Protocol changes that reduce split votes (e.g., ESCAPE~\cite{zhang2022escape}) are complementary to timeout adaptation.
We do not include such baselines because they change protocol behavior and are not directly comparable to election-timeout-only policies; evaluating combinations is an important direction.
\paragraph{Multi-agent non-stationarity.}
Our setting is endogenously non-stationary: each node runs its own online policy, and concurrent exploration changes election contention and thus the data distribution seen by other nodes.
We do not claim formal convergence guarantees in this coupled system.
Instead, we mitigate worst-case behavior with a small action space, a conservative safety wrapper that forces a stable arm after repeated failures, and an update rule that only learns from candidate-election attempts (not from every follower deadline reset), reducing feedback amplification.
We further provide time-series diagnostics (arm fractions, term churn, and safety-entry frequency) and parameter sensitivity analyses in the appendix to assess stability empirically.
Across our evaluated regimes we do not observe persistent harmful synchronization beyond the alignment stress test; safety overlap statistics in \Cref{sec:appendix-safety-overlap} quantify concurrent forced-safe episodes and remain low in the main scenario.
When timer randomization is artificially narrowed, alignment can re-emerge; enforcing a small minimum jitter width is an effective implementation guardrail (\Cref{sec:appendix-alignment}).
Our main scenario also includes a cluster-wide regime switch, and we quantify forced-safe overlap across nodes (\Cref{sec:appendix-safety-overlap}) to gauge correlated behavior under shared shocks.
Future work could explore lightweight coordination to avoid synchronized aggressiveness, such as jittered exploration schedules, per-term randomized exploration coefficients, or cluster-level ``safe-arm'' pins during turbulence.

\subsection{Why this matters for industry}
In practice, election-timeout miscalibration often leads to two costly outcomes: (i) incident-driven ``timeout inflation'' that increases recovery latency for all future failures, and (ii) environment-specific configuration that must be retuned as clusters move across network domains.
BALLAST suggests a path to reduce both costs by making timeouts adapt to locally observed conditions, while keeping the mechanism simple enough to operationalize:
a small number of interpretable arms, microsecond-level inference, and explicit safety gating.
\section{Related Work}
\label{sec:related}

\textbf{Consensus under WAN and liveness failures.}
Wide-area deployments motivate latency-aware or network-flexible consensus designs (e.g., WPaxos~\cite{ailijiang2017wpaxos}) and careful treatment of liveness assumptions.
Recent industrial experience reports emphasize that long-tail delay, recovery turbulence, and failure-detector assumptions can dominate real-world availability behavior~\cite{tennage2024racs}.
BALLAST targets this regime by adapting timeouts online while preserving Raft's safety rules.
\textbf{Analytical models of Raft availability.}
Several works model Raft's performance and availability under timing and loss assumptions (e.g., analytical studies for private blockchains~\cite{huang2018raftanalysis}).
These models complement our empirical simulator by offering closed-form insight into election dynamics and split-vote probabilities; we use them primarily to motivate stress regimes and to interpret timer-alignment effects.

\textbf{Raft election improvements.}
Prior work improves leader election via protocol modifications or additional coordination to reduce split votes and churn.
Examples include explicit split-vote mitigation (ESCAPE)~\cite{zhang2022escape} and various adaptive or multi-strategy election mechanisms~\cite{du2023multistrategy,chen2024adaptiveraftwireless,gou2025draft}.
These approaches often hard-code heuristics or require protocol changes.
BALLAST is complementary: it keeps the Raft protocol intact and learns a policy for the election-timeout \emph{parameter} using lightweight online learning.

\textbf{Dynatune.}
Dynatune dynamically tunes election parameters based on network measurements (RTT and packet loss) obtained from heartbeats, and reports reductions in leader failure detection and out-of-service (OTS) time without altering Raft's core mechanisms~\cite{shiozaki2025dynatune}.
BALLAST differs in two ways: (i) it optimizes an explicit recovery-focused reward rather than relying on a handcrafted mapping from measurements to parameters, and (ii) it focuses on election-timeout selection only (heartbeat interval remains fixed), enabling a smaller and more interpretable action space with explicit safety gating.
We include a simplified co-tuning baseline (\texttt{dynatune\_joint}) to capture the qualitative benefit of joint tuning, but a protocol-faithful Dynatune replication remains future work.

\textbf{Adaptive baselines and non-stationary bandits.}
Beyond UCB-style contextual bandits, recent work studies non-stationary bandit algorithms and discounting/forgetting mechanisms that can be competitive in changing environments.
Thompson sampling~\cite{thompson1933likelihood} and its linear contextual variants~\cite{agrawal2013thompson} provide an alternative exploration mechanism, while non-stationary bandit settings motivate explicit forgetting and stability considerations~\cite{besbes2014nonstationary}.
In parallel, randomized backoff-style mechanisms (e.g., Baxos) and failure-detector adaptivity in MultiPaxos/Raft-style systems motivate simple adaptive baselines that may approximate some benefits of learning without complex models.
We also note recent non-stationary contextual bandits (e.g., discounted or sliding-window LinUCB variants) as plausible alternatives to LinUCB in highly shifting regimes.
We include a discounted Thompson-sampling baseline (\texttt{bandit\_ts\_safe}) and a simple quantile-with-decay adaptive baseline (\texttt{quantile\_decay}) to test whether bandit learning is necessary beyond robust statistics and lightweight adaptivity.
Change-point detection and restart-based bandit variants (e.g., CD-UCB-style frameworks~\cite{liu2017cdmab}) are also relevant for abrupt regime shifts; we leave a head-to-head evaluation to future work.

\textbf{Interference-aware and safe bandits.}
Multi-agent or interference-aware contextual bandit settings study learning under coupling between agents or arms, including distributed/collective linear bandits and bandits with shared feedback structure~\cite{amani2023distributed,korda2016distributed}.
These methods could provide a more principled way to model inter-node interference in leader elections than our independent per-node learners.
Separately, safe exploration for bandits (e.g., conservative or constraint-aware algorithms) provides probabilistic performance guarantees and could replace the current heuristic safety gate with a more formal safety layer~\cite{kazerouni2016conservative,jagerman2020sea}.
Variance-aware UCB variants designed for non-stationary regimes (e.g., RAVEN-UCB) are also plausible alternatives in turbulent environments~\cite{fang2025ravenucb}.
We view these lines of work as complementary and leave their integration to future work.

\textbf{Learning for systems and configuration.}
Learning-based auto-tuning has a long history in systems, including database knob tuning at scale (OtterTune)~\cite{vanaken2017ottertune}.
More recently, online configuration management and adaptive control are increasingly studied in production settings~\cite{harding2025abob}.
In the consensus space, learning has been applied to heavier-weight adaptation in BFT protocols~\cite{wu2025bftbrain}.
BALLAST differs by focusing on a narrow, latency-sensitive control point (election timeouts) and using a white-box contextual bandit (LinUCB) rather than deep RL, enabling microsecond-level inference and transparent safety valves.

\textbf{Positioning.}
Our intent is not to propose a new consensus protocol, but to demonstrate that a small, interpretable learning component can replace brittle manual timeout tuning while respecting the operational constraints of consensus implementations.
This positioning makes BALLAST naturally complementary to protocol-level improvements: even with improved election rules, deployments still face environment shifts where parameter adaptation remains necessary.
\section{Conclusion}
\label{sec:conclusion}

BALLAST replaces static election-timeout heuristics with lightweight contextual bandits and safe exploration.
Across long-tail and recovery-turbulence scenarios, BALLAST reduces recovery time and unwritable time compared to randomized timeouts and common heuristics, while remaining competitive on stable LAN/WAN settings.
These results suggest that white-box online learning can provide practical liveness improvements for consensus protocols without sacrificing deployability.

\paragraph{Practical impact.}
From an operator perspective, BALLAST aims to reduce two kinds of operational pain:
(i) emergency ``timeout inflation'' during incidents, and
(ii) environment-specific tuning when the same software moves across LAN/WAN/cloud settings.
By adapting to local heartbeat and election signals, BALLAST can shift between aggressive and conservative regimes without manual intervention, improving availability under turbulence while preserving steady-state behavior.

\paragraph{Limitations and future work.}
Our evaluation focuses on leader election and availability proxies rather than end-to-end database throughput.
Real implementations also include log replication, disk stalls, and application-level backpressure that can interact with election timing.
Future work should validate BALLAST in a production-quality Raft implementation and explore richer safety constraints (e.g., SLO-aware or cost-aware policies) that explicitly bound worst-case unavailability.
\appendix
\section{Additional Results}
\label{sec:appendix}

This appendix collects additional scenario results that complement the main stress test in \Cref{sec:eval-main}.
We include them for completeness and reproducibility, but we do not rely on them for the paper's primary claims.

\paragraph{Tail metrics (p99 and worst-case).}
\Cref{tab:main-worstcase} reports recovery p99 and worst observed recovery durations (per-seed maxima) for the main scenario, along with worst observed unwritable fractions.
These tail statistics reinforce the trends reported in the main table.

\setcounter{figure}{0}
\setcounter{table}{0}
\renewcommand{\thefigure}{\thesection.\arabic{figure}}
\renewcommand{\thetable}{\thesection.\arabic{table}}
\renewcommand{\theHfigure}{\thesection.\arabic{figure}}
\renewcommand{\theHtable}{\thesection.\arabic{table}}

\begin{table}[htbp]
  \centering
  \scriptsize
  \setlength{\tabcolsep}{3pt}
  \resizebox{\linewidth}{!}{\input{tables/table_main_worstcase}}
  \caption{Main scenario: recovery p99 and worst observed recovery/unwritable metrics (per-seed maxima).}
  \label{tab:main-worstcase}
\end{table}

\subsection{Motivation scenario}
\label{sec:appendix-motivation}
\begin{figure}[htbp]
  \centering
  \includegraphics[width=\linewidth]{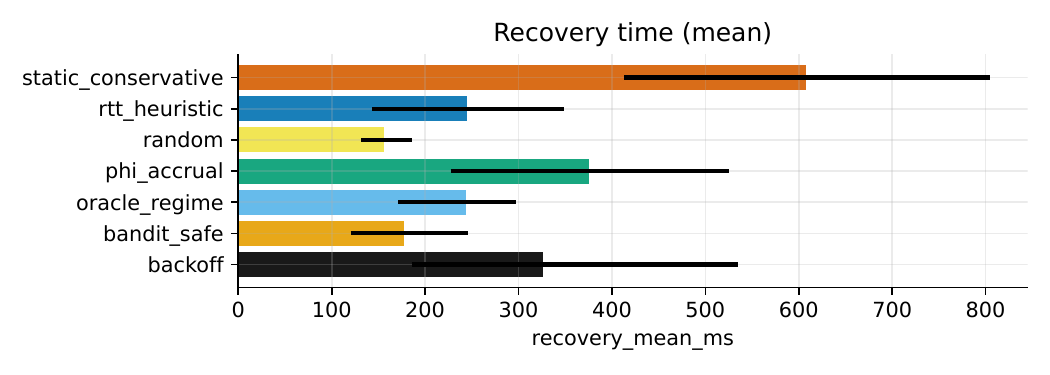}
  \caption{Motivation scenario: mean recovery time with 95\% CI.}
  \label{fig:appendix-motivation-recovery}
\end{figure}

We report the motivation-scenario recovery comparison in \Cref{fig:appendix-motivation-recovery}.

\subsection{Partition recovery turbulence}
\label{sec:appendix-partition}
\begin{figure}[htbp]
  \centering
  \includegraphics[width=\linewidth]{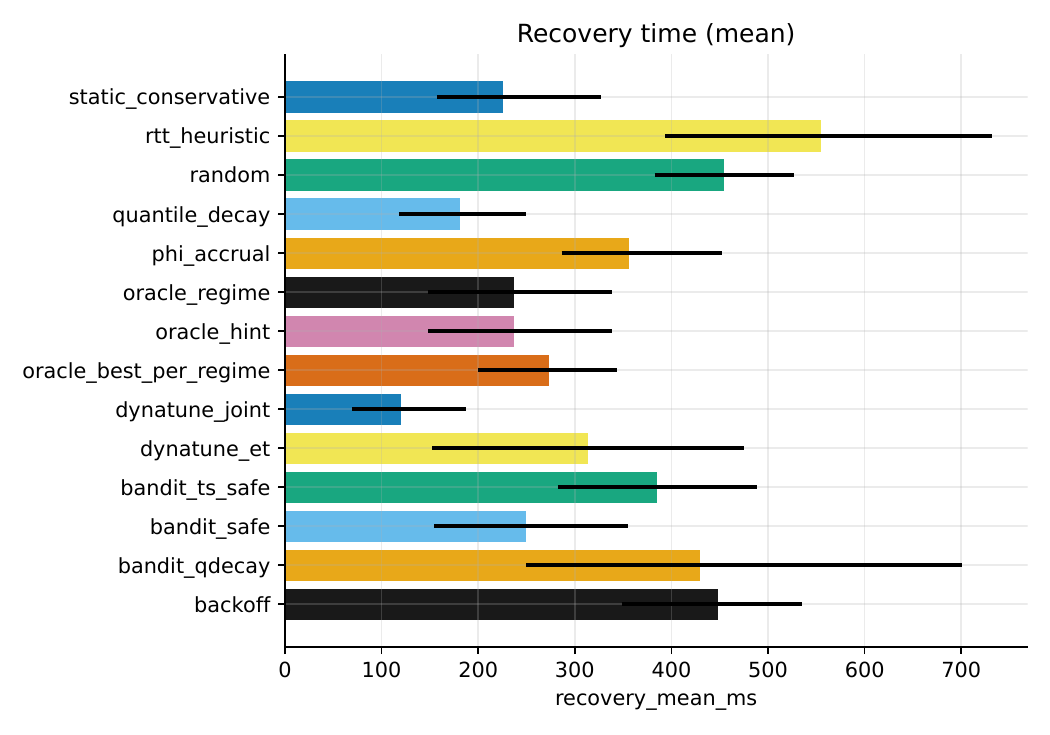}
  \caption{Partition recovery scenario: mean recovery time with 95\% CI.}
  \label{fig:appendix-partition-mean}
\end{figure}

We summarize recovery-time comparisons for the partition-recovery turbulence setting in \Cref{fig:appendix-partition-mean}.

\subsection{Stable WAN and LAN no-regression}
\label{sec:appendix-stable}
\begin{figure}[htbp]
  \centering
  \includegraphics[width=\linewidth]{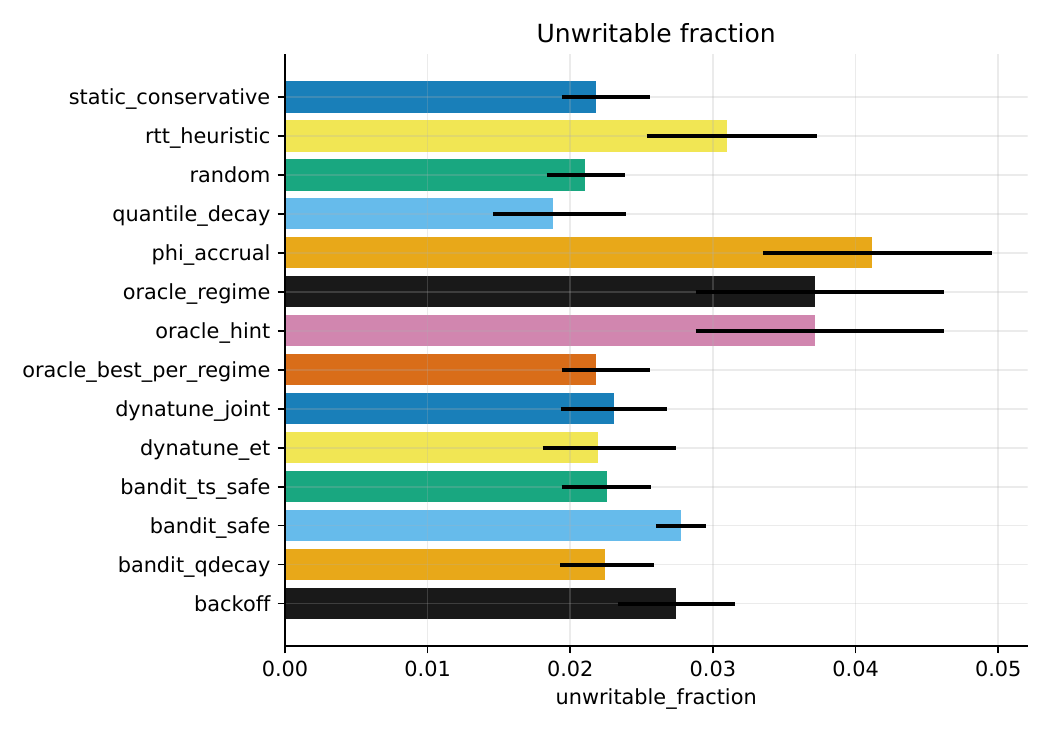}
  \caption{Stable WAN scenario: unwritable fraction with 95\% CI.}
  \label{fig:appendix-wan-unwritable}
\end{figure}

\begin{figure}[htbp]
  \centering
  \includegraphics[width=\linewidth]{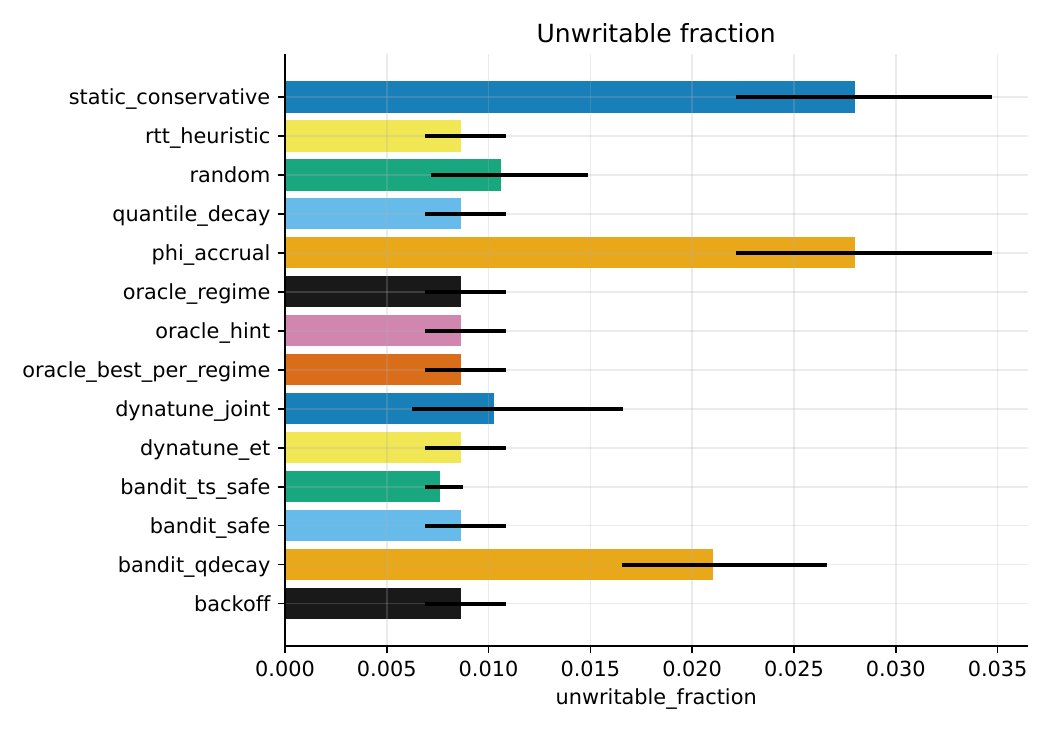}
  \caption{LAN no-regression scenario: unwritable fraction with 95\% CI.}
  \label{fig:appendix-lan-unwritable}
\end{figure}

We report unwritable-fraction comparisons for stable WAN and LAN no-regression in \Cref{fig:appendix-wan-unwritable,fig:appendix-lan-unwritable}.

\subsection{Real-prototype stress tests}
\label{sec:appendix-etcd-stress}
We apply host-level CPU+IO stress during the real \texttt{etcd/raft} prototype runs to emulate GC/IO contention and scheduling noise.
\Cref{tab:etcd-raft-stress,tab:etcd-raft-multi-az-stress} summarize the main-scenario results for the single-host netem setting and the multi-AZ no-netem setting, respectively.
\begin{table}[htbp]
  \centering
  \scriptsize
  \resizebox{\linewidth}{!}{\input{tables/table_etcd_raft_main_stress}}
  \caption{Real \texttt{etcd/raft} prototype under CPU+IO stress (single-host netem): main scenario summary (95\% CI, 10 seeds).}
  \label{tab:etcd-raft-stress}
\end{table}
\begin{table}[htbp]
  \centering
  \scriptsize
  \resizebox{\linewidth}{!}{\input{tables/table_etcd_raft_multi_az_no_netem_stress}}
  \caption{Real \texttt{etcd/raft} prototype under CPU+IO stress (AWS multi-AZ, no \texttt{netem}): main scenario summary (95\% CI, 10 seeds).}
  \label{tab:etcd-raft-multi-az-stress}
\end{table}

\subsection{Real-prototype with PreVote + workload (N=5,7)}
\label{sec:appendix-etcd-prevote}
We extend the Docker+netem prototype to larger clusters with PreVote enabled and a lightweight proposal workload (one proposal every \SI{50}{ms}, 256\,B payload).
\Cref{tab:etcd-raft-main-prevote-n5,tab:etcd-raft-main-prevote-n7} report main-scenario results for $N{=}5$ and $N{=}7$ (5 seeds each).
\begin{table}[htbp]
  \centering
  \scriptsize
  \resizebox{\linewidth}{!}{\input{tables/table_etcd_raft_main_prevote_wl_n5}}
  \caption{Real \texttt{etcd/raft} prototype with PreVote + workload (Docker+netem), $N{=}5$: main scenario summary (95\% CI, 5 seeds).}
  \label{tab:etcd-raft-main-prevote-n5}
\end{table}
\begin{table}[htbp]
  \centering
  \scriptsize
  \resizebox{\linewidth}{!}{\input{tables/table_etcd_raft_main_prevote_wl_n7}}
  \caption{Real \texttt{etcd/raft} prototype with PreVote + workload (Docker+netem), $N{=}7$: main scenario summary (95\% CI, 5 seeds).}
  \label{tab:etcd-raft-main-prevote-n7}
\end{table}

\subsection{Larger clusters (N=15, 21)}
\label{sec:appendix-large-n}
To probe split-vote dynamics at larger scales, we run the main scenario with $N{=}15$ and $N{=}21$ using a core subset of baselines (random, static\_conservative, backoff, quantile\_decay, dynatune\_et, and bandit\_safe) over 10 seeds.
We additionally test two more conservative BALLAST variants without changing the scenario: (i) \texttt{bandit\_qdecay}, which uses quantile-anchored scaling with relative multiplier arms, and (ii) \texttt{bandit\_safe\_ln}, which shifts timeout arms upward and enforces a larger minimum jitter to reduce alignment.
\Cref{tab:appendix-n15,tab:appendix-n21} summarize recovery and unwritable metrics; the BALLAST variants are marked as ``(our)'' and improve large-$N$ stability, with the best choice depending on $N$.
\begin{table}[htbp]
  \centering
  \scriptsize
  \setlength{\tabcolsep}{3pt}
  \resizebox{\linewidth}{!}{\input{tables/table_main_n15_core}}
  \caption{Main scenario with $N{=}15$ (core methods + BALLAST variants, 10 seeds): recovery and unwritable metrics (95\% CI).}
  \label{tab:appendix-n15}
\end{table}

\begin{table}[htbp]
  \centering
  \scriptsize
  \setlength{\tabcolsep}{3pt}
  \resizebox{\linewidth}{!}{\input{tables/table_main_n21_core}}
  \caption{Main scenario with $N{=}21$ (core methods + BALLAST variants, 10 seeds): recovery and unwritable metrics (95\% CI).}
  \label{tab:appendix-n21}
\end{table}

\subsection{Timer-alignment sensitivity}
\label{sec:appendix-alignment}
We stress randomized timer alignment by narrowing each arm to a 1\,ms-wide range around its nominal midpoint, reducing symmetry-breaking randomness.
\Cref{tab:appendix-align} reports recovery, unwritable, and split-vote rates for the core methods (10 seeds).
The narrow ranges amplify split votes, underscoring the importance of randomization even when adaptive policies are used.
\paragraph{Mitigation via minimum jitter.}
We additionally test a simple anti-alignment guard: enforce a minimum jitter width (we use \SI{20}{ms}) when sampling from narrow ranges.
\Cref{tab:appendix-align-jitter} shows that this guard restores stability for \texttt{bandit\_safe} under alignment stress without altering the baseline scenarios, suggesting that a small implementation-level jitter floor is sufficient to avoid this failure mode.
\begin{table}[htbp]
  \centering
  \scriptsize
  \setlength{\tabcolsep}{3pt}
  \resizebox{\linewidth}{!}{\input{tables/table_main_align_narrow}}
  \caption{Main scenario with narrow timeout ranges (alignment stress; core methods, 10 seeds): recovery, unwritable, and split-vote rates.}
  \label{tab:appendix-align}
\end{table}
\begin{table}[htbp]
  \centering
  \scriptsize
  \setlength{\tabcolsep}{3pt}
  \resizebox{\linewidth}{!}{\input{tables/table_main_align_narrow_jitter}}
  \caption{Alignment stress with minimum jitter width (\SI{20}{ms}): core methods, 10 seeds.}
  \label{tab:appendix-align-jitter}
\end{table}

\subsection{Cold-start resets on restart}
\label{sec:appendix-restart}
To emulate process restarts that clear in-memory learning state, we add a reset-on-restart variant in the partition-recovery turbulence scenario.
\Cref{tab:appendix-restart} reports recovery and unwritable metrics for \texttt{bandit\_safe} with and without policy reset on restart.
The safety wrapper mitigates early aggressiveness after reset, and the overall availability degradation is modest relative to the base policy.
\begin{table}[htbp]
  \centering
  \scriptsize
  \setlength{\tabcolsep}{3pt}
  \resizebox{\linewidth}{!}{\input{tables/table_partition_reset_policy}}
  \caption{Partition-recovery turbulence: \texttt{bandit\_safe} with vs.\ without policy reset on restart (5 seeds).}
  \label{tab:appendix-restart}
\end{table}

\subsection{Baseline tuning and parameter sensitivity}
\label{sec:appendix-tuning-sensitivity}
\begin{figure}[htbp]
  \centering
  \includegraphics[width=\linewidth]{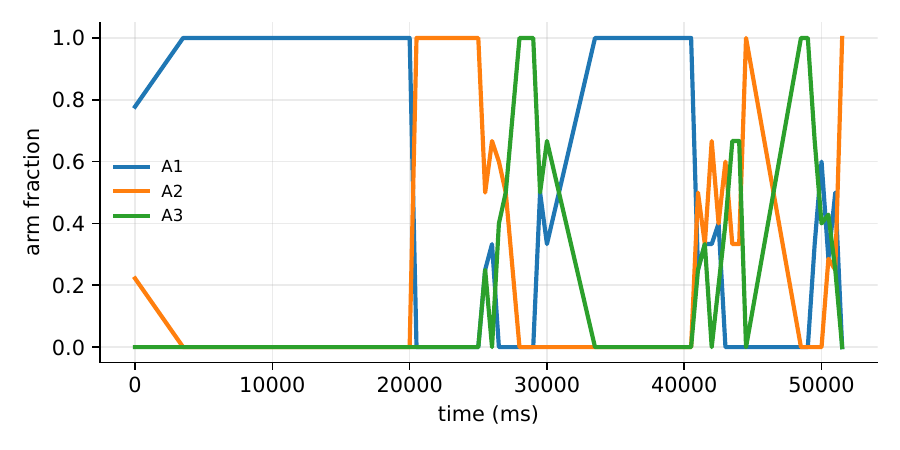}
  \caption{Representative main-scenario run (seed 11): per-bin arm selection fractions for \texttt{bandit\_safe}, illustrating online adaptation under multi-agent interaction.}
  \label{fig:appendix-arm-fraction}
\end{figure}

\begin{figure}[htbp]
  \centering
  \includegraphics[width=\linewidth]{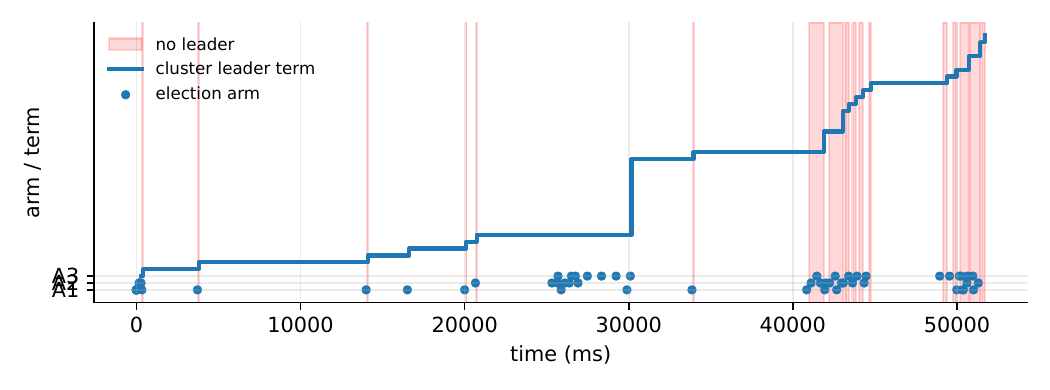}
  \caption{Representative main-scenario run (seed 11): leader term evolution and election arm choices over time.}
  \label{fig:appendix-timeline}
\end{figure}

We summarize tuning parameters and sensitivity results in \Cref{fig:appendix-arm-fraction,fig:appendix-timeline,tab:baseline-tuning,tab:sensitivity}.

\begin{table}[htbp]
  \centering
  \scriptsize
  \setlength{\tabcolsep}{3pt}
  \resizebox{\linewidth}{!}{\input{tables/table_baseline_tuning}}
  \caption{Per-scenario tuned parameters for threshold-based baselines (tuned on disjoint seeds).}
  \label{tab:baseline-tuning}
\end{table}

\begin{table}[htbp]
  \centering
  \scriptsize
  \setlength{\tabcolsep}{3pt}
  \resizebox{\linewidth}{!}{\input{tables/table_sensitivity}}
  \caption{Sensitivity analysis for reward weights, safety parameters, exploration, and feature scaling on the recovery-turbulence scenario (deltas relative to the base configuration).}
  \label{tab:sensitivity}
\end{table}

\subsection{Cross-scenario ablation sanity check}
\label{sec:appendix-ablation-cross}
To support the claim that the base configuration is a cross-scenario compromise, we report deltas for selected ablation variants across the main, partition, WAN-stable, and LAN scenarios.
\Cref{tab:appendix-ablation-cross} reports $\Delta$Recovery/\,$\Delta$Unwritable (percent, relative to base) for each scenario.
\begin{table}[htbp]
  \centering
  \scriptsize
  \setlength{\tabcolsep}{3pt}
  \resizebox{\linewidth}{!}{\input{tables/table_ablation_cross_scenario}}
  \caption{Cross-scenario ablation: deltas vs.\ base (percent). Each cell shows $\Delta$Recovery/\,$\Delta$Unwritable; negative is better.}
  \label{tab:appendix-ablation-cross}
\end{table}

\subsection{Safety cooldown overlap}
\label{sec:appendix-safety-overlap}
We analyze forced-safe cooldown episodes using policy-decision logs and summarize their frequency, mean duration, and overlap across nodes in the main scenario.
\Cref{tab:appendix-safety-overlap} reports averages with 95\% CIs; overlap is the fraction of time with at least two (or three) nodes concurrently in forced-safe mode.
\begin{table}[htbp]
  \centering
  \scriptsize
  \setlength{\tabcolsep}{3pt}
  \resizebox{\linewidth}{!}{\input{tables/table_safety_overlap}}
  \caption{Safety cooldown statistics in the main scenario (30 seeds).}
  \label{tab:appendix-safety-overlap}
\end{table}

\subsection{Failure-cause breakdown for election timeouts}
\label{sec:appendix-failure-breakdown}
To contextualize the split-vote proxy, we classify failed elections into (i) no-quorum-alive, (ii) low reachability (RequestVote observed by fewer than a majority of peers), and (iii) contention-like failures (majority reachable but no leader).
\Cref{tab:appendix-failure-breakdown} shows that no-quorum failures are zero in the main scenario; most failures split between low-reachability and contention-like cases depending on the policy, supporting the use of \texttt{election\_failed} as a churn proxy while acknowledging that some failures stem from reachability loss.
\begin{table}[htbp]
  \centering
  \scriptsize
  \setlength{\tabcolsep}{3pt}
  \resizebox{\linewidth}{!}{\input{tables/table_failure_breakdown}}
  \caption{Main scenario: breakdown of failed elections by likely cause (95\% CI). No-quorum fraction is zero in this scenario and omitted.}
  \label{tab:appendix-failure-breakdown}
\end{table}

\subsection{Action-space sensitivity (broader arms)}
\label{sec:appendix-arms}
We evaluate multiple arm sets to test robustness beyond the 3-arm design in the main body: a broader 5-arm set (up to \SI{2400}{ms}), a shifted 3-arm set that starts at longer ranges, and a finer 7-arm set that adds intermediate ranges.
The safety arm is set to the most conservative arm in each action set.
As shown in \Cref{tab:armset}, shifting all arms upward can reduce unwritable fraction at the cost of slower recovery, while a finer-grained arm set is not uniformly beneficial, underscoring the value of simple, interpretable arm designs and data-driven scaling.
\begin{table}[htbp]
  \centering
  \scriptsize
  \setlength{\tabcolsep}{3pt}
  \resizebox{\linewidth}{!}{\input{tables/table_armset}}
  \caption{Action-space sensitivity on the main scenario: BALLAST under multiple timeout arm sets.}
  \label{tab:armset}
\end{table}

\bibliographystyle{abbrv}
\bibliography{references}

\end{document}

%% file: tables/table_main_methods.tex
\begin{tabular}{lrrrrr}
\toprule
Method & Recovery mean (ms) & Recovery p95 (ms) & Recovery p99 (ms) & Recovery max (ms) & Unwritable frac. \\
\midrule
\texttt{random} & 1100 [927.3, 1257] & 5110 [4217, 5976] & 6937 [5671, 8080] & 7396 [6020, 8626] & 0.3586 [0.3014, 0.4075] \\
\texttt{static\_conservative} & 310.6 [266.2, 364.3] & 1052 [892.2, 1249] & 1189 [982.0, 1446] & 1223 [1006, 1495] & 0.0377 [0.0335, 0.0429] \\
\texttt{backoff} & 442.6 [384.2, 508.2] & 1672 [1490, 1864] & 2316 [2089, 2565] & 2498 [2227, 2800] & 0.1712 [0.1392, 0.2085] \\
\texttt{rtt\_heuristic} & 704.6 [454.7, 984.0] & 2922 [1721, 4318] & 7551 [4556, 10944] & 8710 [5209, 12679] & 0.1817 [0.1210, 0.2487] \\
\texttt{phi\_accrual} & 566.8 [511.4, 623.1] & 2636 [2326, 2945] & 4062 [3620, 4496] & 4460 [3945, 4963] & 0.2345 [0.2080, 0.2618] \\
\texttt{quantile\_decay} & 194.5 [162.9, 231.8] & 714.3 [563.5, 896.1] & 894.0 [687.2, 1153] & 938.9 [717.9, 1218] & 0.0314 [0.0267, 0.0368] \\
\texttt{bandit\_qdecay} & 304.2 [229.8, 406.1] & 1139 [875.6, 1470] & 1341 [1019, 1737] & 1391 [1055, 1805] & 0.0382 [0.0322, 0.0454] \\
\texttt{dynatune\_et} & 226.6 [183.9, 278.7] & 821.5 [656.0, 1031] & 983.6 [777.8, 1241] & 1024 [808.7, 1294] & \textbf{0.0307 [0.0259, 0.0362]} \\
\texttt{dynatune\_joint} & \textbf{105.8 [82.98, 134.0]} & \textbf{416.6 [304.2, 550.0]} & 844.4 [677.4, 1039] & 1032 [822.9, 1262] & 0.0594 [0.0481, 0.0717] \\
\texttt{bandit\_safe} (our) & 153.8 [128.3, 183.3] & 476.4 [366.9, 600.8] & \textbf{659.3 [529.1, 809.2]} & \textbf{705.1 [564.9, 866.2]} & 0.0416 [0.0329, 0.0515] \\
\texttt{bandit\_ts\_safe} & 502.4 [419.5, 581.2] & 1800 [1469, 2134] & 2543 [2013, 3074] & 2728 [2136, 3340] & 0.1594 [0.1286, 0.1904] \\
\texttt{oracle\_hint} & 191.4 [167.8, 217.1] & 592.3 [486.2, 711.0] & 1364 [1061, 1709] & 1576 [1208, 1992] & 0.0704 [0.0611, 0.0805] \\
\texttt{oracle\_best\_per\_regime} & 291.1 [221.7, 372.7] & 1072 [811.7, 1371] & 1365 [1021, 1753] & 1441 [1077, 1861] & 0.0495 [0.0311, 0.0812] \\
\bottomrule
\end{tabular}

%% file: tables/table_etcd_raft_multi_az_no_netem.tex
\begin{tabular}{lcc}
\toprule
Method & Recovery mean (ms, 95\% CI) & Unwritable (mean, 95\% CI)\\
\midrule
random & 2305.6 [2122.2, 2469.5] & 0.2 [0.2, 0.2]\\
static & 2643.9 [2541.7, 2773.7] & 0.2 [0.2, 0.2]\\
backoff & 2308.4 [2194.6, 2459.6] & 0.2 [0.1, 0.2]\\
ballast & 2385.4 [2234.9, 2560.0] & 0.2 [0.2, 0.2]\\
\bottomrule
\end{tabular}

%% file: tables/table_etcd_raft_main.tex
\begin{tabular}{lcc}
\toprule
Method & Recovery mean (ms, 95\% CI) & Unwritable (mean, 95\% CI)\\
\midrule
random & 911.1 [718.5, 1156.4] & 0.2 [0.1, 0.2]\\
static & 1292.9 [815.4, 1858.4] & 0.2 [0.1, 0.2]\\
backoff & 680.5 [551.7, 804.6] & 0.1 [0.1, 0.1]\\
ballast & 588.1 [521.2, 664.8] & 0.1 [0.0, 0.1]\\
\bottomrule
\end{tabular}

%% file: tables/table_overhead.tex
\begin{tabular}{lrrrr}
\toprule
Operation & Mean (\si{\micro\second}) & p50 & p95 & p99 \\
\midrule
ChooseArm & 24.97 & 24.50 & 26.46 & 45.88 \\
Update & 3.72 & 3.58 & 3.92 & 5.62 \\
\bottomrule
\end{tabular}

%% file: tables/table_etcd_overhead.tex
\begin{tabular}{lrrrrr}
\toprule
Operation & Mean (\si{\micro\second}) & p50 & p95 & p99 & Max \\
\midrule
ChooseArm & 14.34 & 4.00 & 31.00 & 71.00 & 22297 \\
Update & 1.68 & 1.00 & 4.00 & 6.00 & 17.00 \\
\bottomrule
\end{tabular}

%% file: tables/table_ablation.tex
\begin{tabular}{lrrr}
\toprule
Ablation & $\Delta$Recovery mean (\%) & $\Delta$Unwritable (\%) & $\Delta$Split-vote rate (\%) \\
\midrule
\texttt{base} & 0.0 & 0.0 & 0.0 \\
\texttt{no\_safe\_exploration} & -7.9 & -6.2 & 22.55 \\
\texttt{nonstationary\_sliding} & -49.92 & -41.37 & -17.96 \\
\texttt{explore\_low\_alpha} & -37.75 & -29.61 & 1.3 \\
\texttt{explore\_high\_alpha} & -37.10 & -27.40 & -10.37 \\
\texttt{reward\_success\_only} & -29.77 & -18.46 & -7.7 \\
\texttt{ctx\_hb\_only} & 3.9 & 3.2 & -0.7 \\
\bottomrule
\end{tabular}

%% file: tables/table_main_worstcase.tex
\begin{tabular}{lrrr}
\toprule
Method & Recovery p99 (max over seeds) & Recovery max (max over seeds) & Unwritable frac. (max over seeds) \\
\midrule
\texttt{random} & 12667 & 13700 & 0.4769 \\
\texttt{static\_conservative} & 3388 & 3521 & 0.0788 \\
\texttt{backoff} & 4377 & 4921 & 0.3869 \\
\texttt{rtt\_heuristic} & 22082 & 24671 & 0.4521 \\
\texttt{phi\_accrual} & 6237 & 7025 & 0.3873 \\
\texttt{quantile\_decay} & 3395 & 3624 & 0.0697 \\
\texttt{bandit\_qdecay} & 4710 & 4837 & 0.0939 \\
\texttt{dynatune\_et} & 3479 & 3682 & 0.0835 \\
\texttt{dynatune\_joint} & 2253 & 2424 & 0.1469 \\
\texttt{bandit\_safe} (our) & 2020 & 2163 & 0.1235 \\
\texttt{bandit\_ts\_safe} & 5437 & 6310 & 0.3150 \\
\texttt{oracle\_hint} & 3763 & 4645 & 0.1320 \\
\texttt{oracle\_best\_per\_regime} & 4515 & 4898 & 0.4286 \\
\bottomrule
\end{tabular}

%% file: tables/table_etcd_raft_main_stress.tex
\begin{tabular}{lcc}
\toprule
Method & Recovery mean (ms, 95\% CI) & Unwritable (mean, 95\% CI)\\
\midrule
random & 1966.7 [1678.6, 2352.6] & 0.2 [0.2, 0.3]\\
static & 2631.4 [2396.1, 2838.3] & 0.2 [0.2, 0.2]\\
backoff & 2098.6 [1889.9, 2337.4] & 0.2 [0.2, 0.2]\\
ballast & 2056.4 [1823.9, 2266.5] & 0.2 [0.2, 0.2]\\
\bottomrule
\end{tabular}

%% file: tables/table_etcd_raft_multi_az_no_netem_stress.tex
\begin{tabular}{lcc}
\toprule
Method & Recovery mean (ms, 95\% CI) & Unwritable (mean, 95\% CI)\\
\midrule
random & 2254.4 [2078.9, 2409.8] & 0.2 [0.2, 0.2]\\
static & 2750.6 [2627.8, 2874.4] & 0.2 [0.2, 0.2]\\
backoff & 2321.5 [2186.2, 2443.9] & 0.2 [0.2, 0.2]\\
ballast & 2353.2 [2285.8, 2414.6] & 0.2 [0.2, 0.2]\\
\bottomrule
\end{tabular}

%% file: tables/table_etcd_raft_main_prevote_wl_n5.tex
\begin{tabular}{lcc}
\toprule
Method & Recovery mean (ms, 95\% CI) & Unwritable (mean, 95\% CI)\\
\midrule
random & 729.5 [505.5, 1059.1] & 0.039 [0.019, 0.058]\\
static & 756.7 [530.3, 1042.8] & 0.047 [0.029, 0.068]\\
backoff & 505.5 [283.2, 727.8] & 0.041 [0.026, 0.056]\\
ballast & 609.1 [372.4, 847.7] & 0.050 [0.031, 0.069]\\
\bottomrule
\end{tabular}

%% file: tables/table_etcd_raft_main_prevote_wl_n7.tex
\begin{tabular}{lcc}
\toprule
Method & Recovery mean (ms, 95\% CI) & Unwritable (mean, 95\% CI)\\
\midrule
random & 591.6 [439.0, 802.1] & 0.035 [0.017, 0.061]\\
static & 918.3 [464.8, 1666.1] & 0.071 [0.052, 0.098]\\
backoff & 341.8 [245.7, 478.1] & 0.037 [0.021, 0.056]\\
ballast & 539.3 [324.4, 766.6] & 0.042 [0.030, 0.062]\\
\bottomrule
\end{tabular}

%% file: tables/table_main_n15_core.tex
\begin{tabular}{lrrr}
\toprule
Method & Recovery mean (ms) & Recovery p95 (ms) & Unwritable frac. \\
\midrule
\texttt{random} & 1419 [1209, 1626] & 5894 & 0.4425 [0.4080, 0.4692] \\
\texttt{static\_conservative} & 237.0 [172.8, 333.1] & 1030 & 0.0432 [0.0349, 0.0541] \\
\texttt{backoff} & 415.0 [341.2, 483.2] & 1828 & 0.2063 [0.1534, 0.2583] \\
\texttt{quantile\_decay} & 311.5 [218.6, 446.9] & 1262 & 0.0490 [0.0402, 0.0583] \\
\texttt{dynatune\_et} & 134.9 [104.1, 176.0] & 517.7 & 0.0289 [0.0217, 0.0368] \\
\texttt{bandit\_safe (our)} & 224.7 [180.8, 274.2] & 882.7 & 0.0872 [0.0614, 0.1215] \\
\texttt{bandit\_qdecay (our)} & 462.6 [276.8, 692.3] & 1861 & 0.0617 [0.0453, 0.0797] \\
\texttt{bandit\_safe\_ln (our)} & 190.5 [124.4, 266.8] & 699.9 & 0.0348 [0.0256, 0.0465] \\
\bottomrule
\end{tabular}

%% file: tables/table_main_n21_core.tex
\begin{tabular}{lrrr}
\toprule
Method & Recovery mean (ms) & Recovery p95 (ms) & Unwritable frac. \\
\midrule
\texttt{random} & 1218 [1093, 1378] & 2582 & 0.4756 [0.4625, 0.4878] \\
\texttt{static\_conservative} & 154.9 [121.1, 192.9] & 597.8 & 0.0328 [0.0273, 0.0391] \\
\texttt{backoff} & 373.2 [297.3, 443.8] & 1741 & 0.2199 [0.1625, 0.2748] \\
\texttt{quantile\_decay} & 175.5 [111.4, 278.6] & 687.2 & 0.0292 [0.0185, 0.0460] \\
\texttt{dynatune\_et} & 181.7 [132.8, 244.7] & 738.3 & 0.0440 [0.0304, 0.0600] \\
\texttt{bandit\_safe (our)} & 285.0 [218.8, 361.5] & 1084 & 0.1570 [0.1141, 0.2012] \\
\texttt{bandit\_qdecay (our)} & 175.0 [130.3, 226.5] & 720.6 & 0.0331 [0.0267, 0.0404] \\
\texttt{bandit\_safe\_ln (our)} & 180.0 [140.5, 221.4] & 703.6 & 0.0432 [0.0335, 0.0548] \\
\bottomrule
\end{tabular}

%% file: tables/table_main_align_narrow.tex
\begin{tabular}{lrrr}
\toprule
Method & Recovery mean (ms) & Unwritable frac. & Split-vote rate \\
\midrule
\texttt{random} & 727.9 [341.7, 1178] & 0.3621 & 0.1816 \\
\texttt{static\_conservative} & 60000 [60000, 60000] & 0.9058 & 1.0000 \\
\texttt{backoff} & 24506 [6735, 42310] & 0.7204 & 0.4612 \\
\texttt{quantile\_decay} & \textbf{132.1 [105.0, 165.6]} & \textbf{0.2402} & \textbf{0.0237} \\
\texttt{dynatune\_et} & 264.7 [142.2, 434.0] & 0.2752 & 0.0292 \\
\texttt{bandit\_safe} (our) & 24298 [6435, 42157] & 0.6790 & 0.4466 \\
\bottomrule
\end{tabular}

%% file: tables/table_main_align_narrow_jitter.tex
\begin{tabular}{lrrr}
\toprule
Method & Recovery mean (ms) & Unwritable frac. & Split-vote rate \\
\midrule
\texttt{random} & 431.4 [166.0, 783.7] & 0.3243 & 0.1125 \\
\texttt{static\_conservative} & 35779 [19961, 51620] & 0.8707 & 0.7892 \\
\texttt{backoff} & 248.4 [167.9, 338.8] & 0.4128 & 0.0559 \\
\texttt{quantile\_decay} & \textbf{132.1 [105.0, 165.6]} & \textbf{0.2402} & \textbf{0.0237} \\
\texttt{dynatune\_et} & 302.2 [184.4, 462.5] & 0.2518 & 0.0339 \\
\texttt{bandit\_safe} (our) & 238.5 [167.2, 319.6] & 0.4062 & 0.0464 \\
\bottomrule
\end{tabular}

%% file: tables/table_partition_reset_policy.tex
\begin{tabular}{lrrrr}
\toprule
Restart policy & Recovery mean (ms) & Recovery p95 (ms) & Recovery p99 (ms) & Unwritable frac. \\
\midrule
\texttt{no\_reset} & 249.7 [154.3, 354.9] & 909.0 & 1149 & 0.0744 [0.0517, 0.0971] \\
\texttt{reset\_on\_restart} & 244.3 [147.5, 345.3] & 878.8 & 1182 & 0.0695 [0.0426, 0.0965] \\
\bottomrule
\end{tabular}

%% file: tables/table_baseline_tuning.tex
\begin{tabular}{|l|l|}
\hline
Baseline & Tuned parameters \\
\hline
\texttt{rtt\_heuristic} & \makecell[l]{\texttt{thr=[50, 200]}} \\
\hline
\texttt{phi\_accrual} & \makecell[l]{\texttt{phi=[2.0, 3.0]}} \\
\hline
\texttt{dynatune\_et} & \makecell[l]{\texttt{clamp\_max\_ms=2400} \\ \texttt{min\_hb\_ratio=5.0} \\ \texttt{oneway\_factor=2.0} \\ \texttt{safety\_factor=4.0}} \\
\hline
\texttt{dynatune\_joint} & \makecell[l]{\texttt{clamp\_max\_ms=2400} \\ \texttt{hb\_max\_ms=1000} \\ \texttt{hb\_min\_ms=20} \\ \texttt{heartbeat\_ratio=4.0} \\ \texttt{min\_hb\_ratio=4.0} \\ \texttt{oneway\_factor=2.0} \\ \texttt{safety\_factor=2.0}} \\
\hline
\texttt{quantile\_decay} & \makecell[l]{\texttt{p=0.9} \\ \texttt{mult=[3.0,10.0]}} \\
\hline
\texttt{bandit\_qdecay} & \makecell[l]{\texttt{p=0.95} \\ \texttt{mult~[3.0,9.0]}} \\
\hline
\texttt{oracle\_best\_per\_regime} & \makecell[l]{\texttt{map=0->A2; 1->A2}} \\
\hline
\texttt{bandit\_ts\_safe} & \makecell[l]{\texttt{ts\textbackslash{}\_scale=1.5}} \\
\hline
\end{tabular}

%% file: tables/table_sensitivity.tex
\begin{tabular}{lrrr}
\toprule
Ablation & $\Delta$Recovery mean (\%) & $\Delta$Unwritable (\%) & $\Delta$Split-vote rate (\%) \\
\midrule
\texttt{base} & 0.0 & 0.0 & 0.0 \\
\texttt{discount\_0.95} & 13.29 & 16.47 & 2.2 \\
\texttt{discount\_0.995} & -15.47 & -22.04 & 11.98 \\
\texttt{explore\_alpha\_0.2} & 0.5 & -11.43 & 6.7 \\
\texttt{explore\_alpha\_2.0} & 4.7 & 10.0 & 8.0 \\
\texttt{feat\_norm\_z} & -1.6 & 1.7 & 6.3 \\
\texttt{feat\_norm\_z\_clip3} & -3.2 & -4.6 & 0.2 \\
\texttt{reward\_beta\_0} & 13.56 & 7.5 & 3.8 \\
\texttt{reward\_beta\_0.001} & 7.1 & 3.2 & 3.0 \\
\texttt{reward\_beta\_0.004} & -14.74 & -8.5 & -2.9 \\
\texttt{reward\_gamma\_0} & 10.0 & 11.03 & 4.8 \\
\texttt{reward\_gamma\_0.5} & 3.3 & 7.8 & 4.9 \\
\texttt{reward\_gamma\_2} & -17.30 & -10.63 & 7.7 \\
\texttt{safe\_F\_2} & -0.5 & -7.6 & -7.8 \\
\texttt{safe\_F\_4} & -5.9 & -6.5 & 0.7 \\
\texttt{safe\_cd\_1} & 0.1 & 3.8 & 1.0 \\
\texttt{safe\_cd\_4} & 0.3 & -0.1 & 0.1 \\
\texttt{window\_100} & -27.41 & -25.00 & 7.0 \\
\texttt{window\_400} & -27.41 & -25.00 & 7.0 \\
\bottomrule
\end{tabular}

%% file: tables/table_ablation_cross_scenario.tex
\begin{tabular}{lrrrr}
\toprule
Variant & Main & Partition & WAN & LAN \\
\midrule
\texttt{nonstationary\_sliding} & 10.0\% / -1.9\% & 2.9\% / -10.66\% & 12.27\% / -26.54\% & 0.0\% / 0.0\% \\
\texttt{explore\_high\_alpha} & -17.93\% / -19.35\% & 22.49\% / 37.70\% & -8.2\% / 10.31\% & 0.0\% / 0.0\% \\
\texttt{reward\_success\_only} & -17.93\% / -19.35\% & 20.44\% / 38.45\% & -8.4\% / 11.19\% & 0.0\% / 0.0\% \\
\bottomrule
\end{tabular}

%% file: tables/table_safety_overlap.tex
\begin{tabular}{lrrrr}
\toprule
Method & Episodes & Mean dur. (ms) & Overlap $\geq$2 & Overlap $\geq$3 \\
\midrule
\texttt{bandit\_safe} & 1.23 [0.90, 1.57] & 22507 [17233, 27126] & 0.215 [0.129, 0.309] & 0.033 [0.000, 0.083] \\
\texttt{bandit\_ts\_safe} & 4.37 [3.57, 5.10] & 17036 [13624, 20288] & 0.301 [0.237, 0.367] & 0.256 [0.200, 0.310] \\
\bottomrule
\end{tabular}

%% file: tables/table_failure_breakdown.tex
\begin{tabular}{lrr}
\toprule
Method & Low-reach frac. & Contention frac. \\
\midrule
\texttt{random} & 0.3151 [0.2608, 0.3773] & 0.6849 [0.6227, 0.7392] \\
\texttt{quantile\_decay} & 0.5122 [0.4079, 0.6101] & 0.4878 [0.3899, 0.5921] \\
\texttt{bandit\_qdecay} & 0.4076 [0.3124, 0.5014] & 0.5924 [0.4986, 0.6876] \\
\texttt{bandit\_safe} & 0.6979 [0.6159, 0.7690] & 0.3021 [0.2310, 0.3841] \\
\bottomrule
\end{tabular}

%% file: tables/table_armset.tex
\begin{tabular}{lrrr}
\toprule
Setting & Recovery mean (ms) & Unwritable frac. & Split-vote rate \\
\midrule
\texttt{BALLAST (3 arms)} & 153.8 [128.3, 183.3] & 0.0416 [0.0329, 0.0515] & 0.2761 [0.2534, 0.2978] \\
\texttt{BALLAST (5 arms)} & 150.8 [130.4, 171.5] & 0.0415 [0.0346, 0.0494] & 0.2640 [0.2421, 0.2858] \\
\texttt{BALLAST (3 arms shifted)} & 185.1 [151.1, 226.0] & 0.0297 [0.0247, 0.0361] & 0.2361 [0.2185, 0.2558] \\
\texttt{BALLAST (7 arms fine)} & 185.6 [162.0, 211.6] & 0.0558 [0.0477, 0.0646] & 0.3018 [0.2809, 0.3221] \\
\bottomrule
\end{tabular}